\documentclass[a4paper,conference]{IEEEtran}
\IEEEoverridecommandlockouts
\usepackage{cite}
\usepackage{amsmath,amssymb,amsfonts}
\usepackage{algorithmic}
\usepackage{graphicx}
\usepackage{textcomp}
\usepackage{xcolor}
\usepackage[breaklinks=true,bookmarks=false]{hyperref}
\hypersetup{
  colorlinks   = true, %Colours links instead of ugly boxes
  urlcolor     = blue, %Colour for external hyperlinks
  linkcolor    = blue, %Colour of internal links
  citecolor   = red %Colour of citations
}
\usepackage{booktabs}
\usepackage{multirow}
\usepackage{float}
\usepackage[hypcap=false]{caption}
\usepackage{graphicx}
\usepackage[section]{placeins}
\usepackage{etoolbox,lipsum}
\usepackage{afterpage}
\usepackage{booktabs}
\usepackage{subcaption}
\usepackage{capt-of,etoolbox}
\usepackage[section]{placeins}
\usepackage{etoolbox,lipsum}
\usepackage{multirow, makecell}
\usepackage{multirow}
\usepackage{xcolor,colortbl}
\usepackage{soul}
\usepackage{color}
\newcommand{\mypar}[1]{\vspace{0.3cm}\noindent\textbf{#1}}
\usepackage{expl3}
\ExplSyntaxOn
\newcommand\latinabbrev[1]{
  \peek_meaning:NTF . {% Same as \@ifnextchar
    #1\@}%
  { \peek_catcode:NTF a {% Check whether next char has same catcode as \'a, i.e., is a letter
      #1.\@ }%
    {#1.\@}}}
\ExplSyntaxOff
\def\eg{\latinabbrev{e.g}}
\def\etal{\latinabbrev{et al}}

\def\ie{\latinabbrev{i.e}}

\usepackage{tikz}
\newcommand*\circled[1]{\tikz[baseline=(char.base)]{
		\node[shape=circle,draw,inner sep=0.2pt] (char) {#1};}}

\usepackage{balance}
\usepackage{stfloats}
\definecolor{lightsteelblue}{RGB}{176,196,222}
\definecolor{lightsteelred}{RGB}{230,176,160}
\definecolor{lightsteellila}{RGB}{175,181,224}
\definecolor{lightsteelgreen}{RGB}{182,214,207}
\definecolor{lightsteelyellow}{RGB}{240,240,160}
\definecolor{lightsteelwhite}{RGB}{255,255,255}
\definecolor{lightsteelgray}{RGB}{205,201,201}
\definecolor{lightsteellightgray}{RGB}{210,210,210}

\definecolor{pp_blue}{RGB}{68,114,196}
\definecolor{pp_orange}{RGB}{237,125,49}
\definecolor{pp_lila}{RGB}{112,48,160}
\definecolor{pp_ygreen}{RGB}{112,173,71}

\definecolor{gg_blue}{RGB}{52,138,189}
\definecolor{gg_red}{RGB}{226,74,51}
\definecolor{gg_lila}{RGB}{152,142,213}
\definecolor{gg_green}{RGB}{112,173,71}

\definecolor{table_standard}{RGB}{230,153,0}
\definecolor{table_uncertainty}{RGB}{112,48,160}

\usepackage{tikz}

\begin{document}
% hier mehrere Titelversionen:
\title{Uncertainty-sensitive Activity Recognition: \\ A  Reliability Benchmark and the CARING Models}

\author{
Alina Roitberg  \quad \quad  Monica Haurilet \quad \quad  Manuel Martinez\quad \quad  Rainer Stiefelhagen 
\\
\\Institute for Anthropomatics and Robotics
\\ Karlsruhe Institute of Technology, Germany
\\  {\tt\small \{firstname.lastname\}@kit.edu}
}
\maketitle

		\begin{abstract}
	
		Beyond assigning the correct  class, an activity recognition model should also be able to determine, how certain it is in its predictions.  
		We present the first study of how well the confidence values of modern action recognition architectures indeed reflect the probability of the correct outcome and propose a learning-based approach for improving it. 
		First, we extend two popular action recognition datasets with a \emph{reliability} benchmark in form of the expected calibration error and reliability diagrams. 
		Since our evaluation highlights that  confidence values of standard action recognition architectures do not represent the uncertainty well, we introduce a new approach which learns to transform the model output into realistic confidence estimates through an additional calibration network.
		The main idea of our Calibrated Action Recognition with Input Guidance (CARING) model is to learn an optimal scaling parameter \emph{depending on the video representation}. 
		We compare our model with the native action recognition networks and the temperature scaling approach - a wide spread calibration method utilized in image classification.
		While temperature scaling alone drastically improves the reliability of the confidence values, our CARING method consistently leads to the best uncertainty estimates in all benchmark settings.%, leading to the expected calibration  of around $5\%$ on Drive\&Act.
	\end{abstract}
	
	%END SHORT ABSTRACT VERSION

%%%%%%%%% BODY TEXT

%!TEX root = ../root.tex
\section{Introduction}

Humans have a natural grasp of probabilities~\cite{fontanari2014probabilistic}: 
If we hear that a certain event is detected in a video by a neural network with $99$\% confidence, we automatically assume this to be the case. % to be the case oder to be true... to be the truth find ich nicht so gut
%If we hear that a a certain event is deceted in a video with 99\% confidence, we automatically see this as the truth.
\begin{figure}
	\centering
	\includegraphics[width=\linewidth,trim={0 0 0.29cm 0.0cm},clip]{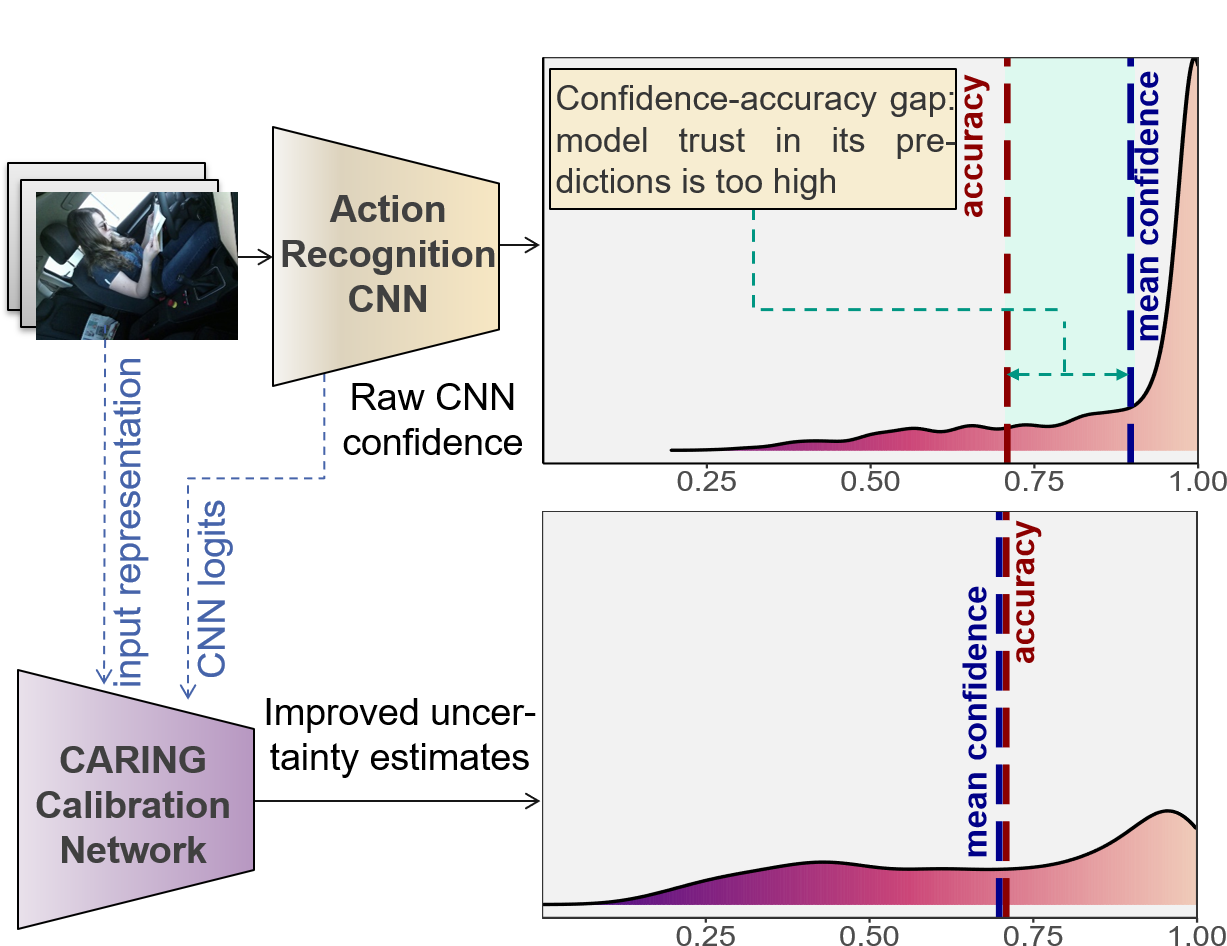}
	\vspace{0.1cm}
	\caption{Softmax confidence distribution of a popular video classification network (Pseudo 3D ResNet tested on a  Drive\&Act validation split) before and after the improvement through our \textbf{C}alibrated\textbf{ A}ction\textbf{ R}ecognition with \textbf{In}put \textbf{G}uidance model.
		Native confidence values underestimate model uncertainty (the majority of samples was rated with $>90\%$ confidence, while the accuracy is significantly lower).
		We propose to incorporate the \emph{reliability} of model confidence in the evaluation of activity recognition models and develop algorithms for improving it.
	}
\label{fig:intro}
\vspace{-0.45cm}
\end{figure}
%Such assumption however would be naive, as the inference merely gives us values of the last fully-connected layer which are usually optimized for high top-1 accuracy on a static set of previously defined actions.
Such assumption however would be naive, as the inference merely gives us values of the last fully-connected layer which are usually optimized for a high top-1 accuracy on a fixed set of previously defined categories.
As these values are usually normalized through the Softmax function to sum up to one, they \emph{appear} to be class probabilities but they do not depict the true confidence of the model~\cite{gal2016dropout}. 
Besides, when engineers apply such deep learning models in practice, they will quickly discover the phenomenon of \emph{model miscalibration}~\ie~the resulting Softmax scores tend to be biased towards very high values~\cite{guo2017calibration,gal2016dropout}.
Unfortunately, such high confidence values are not only present in correct predictions but also in case of misclassifications. 
%Despite impressive results in conventional classification, such overly self-confident models become a burden in applications, where assesing model uncertainty in its prediction plays a central role.
%Despite impressive results in conventional classification, such overly self-confident models become a burden in applications, and might lead to tragic outcomes if assessing model uncertainty in its prediction plays an important role\footnote{A Tragic Loss | Tesla Inc. June 30, 2016. Accessed February 28, 2020. \hyperref[]{https://www.tesla.com/blog/tragic-loss}. }.
Despite impressive results in conventional classification, such overly self-confident models become a burden in applications, and might lead to tragic outcomes if assessing model uncertainty in its prediction plays an important role.
Apart from the direct benefits of proper confidence values for decision-making systems, good assessment of uncertainty enhances model interpretability. 
For example, in the realistic scenario of open-world recognition, low-confidence input might be passed to human experts, which would provide the correct annotations (\ie active learning) and therefore improve the decision boundary. \looseness=-1 

%where assesing model uncertainty in its prediction plays a central role.
%The resulting scores are far away from reflecting true probabilities

%When engineers apply their deep activity recognition models in practice, they are very likely to face the phenomenon of model miscalibration. Traditionally, action recognition algorithms focus on maximizing the top-1 classification performance on a static set of actions. The output of the last fully-connected layer is normalized using the Softmax function, resulting in point estimates from which Cross-Entropy loss is computed. The resulting Softmax scores are often inaccurately denoted as class probabilities and tend to be biased towards very high values. Unfortunately, such high confidence values are not only present for correctly predicted samples but also in case of misclassification or uncertainty. While giving excellent results in closed-set classification, such overly self-confident models become a burden under open-set conditions and in safety-critical applications.

Uncertainty-aware models are vital for safety-critical applications of \emph{activity recognition} approaches, which range from robotics and manufacturing to autonomous driving and surveillance~\cite{roitberg2015multimodal,gebert2019end,Roitberg2020_InterpretableCNN}. 
%The impressive gain in top-1 recognition rate on conventional action recognition datasets linked to the rise of deep learning may draw a rather idealistic picture, due to optimization on a \emph{static} set of carefully designed actions~\cite{carreira2017quo}. 
%The impressive progress reported on the conventional action recognition datasets linked to the rise of deep learning may draw a rather idealistic picture, due to optimization on a \emph{static} set of carefully designed actions~\cite{carreira2017quo}. 
%While recent research in the area of image classification underlines the importance of addressing miscalibration of model confidence, this preformance aspect did not recieve any attention in the field of activity recognition yet.  
While obtaining well-calibrated probability estimates is a growing area in general image recognition~\cite{guo2017calibration,hendrycks17baseline}, this performance aspect did not yet receive any attention in the field of video classification.
%considering the calibration of model confidence scores
The impressive progress reported on the conventional action recognition benchmarks linked to the rise of deep learning~\cite{carreira2017quo,qiu2017learning,hara2018can} may therefore draw a rather idealistic picture, as their validation is often limited to the top-1 accuracy on a static set of carefully designed actions~\cite{kuehne2013hmdb51,carreira2017quo,martin2019drive}. 
While such neural networks are notably bad at detecting data ambiguities, examining how well the confidence values of activity recognition models indeed reflect the probability of a correct prediction has been overlooked in the past and is the main motivation of our work.

In this paper, we aim to elevate the role of uncertainty in the field of activity recognition and develop models, which do not only select the correct behavior class but are also able to \emph{identify misclassifications}.
In other words, the resulting probability value should indeed reflect the likelihood of the prediction to be correct.
To this intent, we propose to incorporate the \emph{reliability} of model confidence in the evaluation of activity recognition models and
develop methods which transform oftentimes biased confidence outputs of the native action recognition models into reliable probability estimates.

 \mypar{Contributions and Summary}
 We argue, that for applications in industrial systems, activity recognition models must not only be accurate, but should also asses, how likely they are to be correct in their prediction through realistic confidence values. 
 %We  therefore go beyond the traditional goal of high top-1 accuracy and make the first step towards activity recognition models capable of \emph{identifying their misclassifications}.
 %This paper therefore goes beyond the traditional goal of high top-1 accuracy and make the first step towards activity recognition models capable of \emph{identifying their misclassifications}.
  This paper  makes the first step towards activity recognition models capable of \emph{identifying their misclassifications} and has thee major contributions.
  %behind spatiotemporal CNNs for driver monitoring, and
%This works aims bring the existing CNN-based approaches for activity recognition closer real-life applications of driver monitoring under presence of uncertainty. The main sources of model uncertainty are scarce training data and dynamic open-set environment. To achieve this, we develop uncertainty-aware models which are able to recognize previously known activities, asses their uncertainty (i.e. identify the cases of misclassification or detect novel classes), and find a way to handle such uncertain examples. 
(1) We present the first  study of how well  the confidence of the modern activity recognition architectures indeed reflects the likelihood of a prediction being correct. 
To this intent, we incorporate the Expected Calibration Error metric in the evaluation procedure of two action recognition CNNs: Pseudo 3D ResNet (P3D)~\cite{qiu2017learning} and Inflated 3D ConvNet (I3D)~\cite{carreira2017quo}. 
Our experiments on two action recognition datasets confirm, that the out-of-the-box probability values of such models do not reflect model uncertainty well (\eg~over $20\%$ expected calibration error on HMDB-51~\cite{kuehne2013hmdb51}).
%Although modern action recognition models deliver excellent recognition rates on static dataset \cite{}, thay are notably bad at detecting data ambiguities.
 %We further aim for a framework which learns to produce realistic confidence estimates and explore two ways to implement our idea.
 %(2) We further aim for a framework which learns to transform the poorly calibrated confidence values of the native action recognition models into more realistic probability estimates and explore two ways to implement our idea. 
 %We first enhance action recognition models with temperature scaling method, a prominent approach for model calibration in image recognition, which learns a single temperature parameter $T$ used to scale the network logits. 
 (2) We further aim for a framework which learns to transform the poorly calibrated confidence values of the native action recognition models into more realistic probability estimates.
 We enhance these architecture with the temperature scaling method~\cite{guo2017calibration}, a prominent approach for model calibration in image recognition, which learns a single temperature parameter $T$ used to scale the network logits. 
 %This method, however, learns \emph{one global temperature value} for scaling,~\ie~after calibration, the logit values are divided by the same scalar independent of the input.
 This method, however, learns \emph{one global temperature value} for scaling,~\ie~after calibration, the logit values are always divided by the same scalar.
 (3) We believe, that input representation gives us significant cues for quantifying network uncertainty, and present a new method for \textbf{C}alibrated \textbf{A}ction \textbf{R}ecognition with \emph{\textbf{In}put \textbf{G}uidance} (CARING). 
 In contrast to~\cite{guo2017calibration}, CARING entails an additional calibration network, which takes as input intermediate representations of the observed activity and learns to produce  \emph{temperature values specific to this input}.  
 While temperature scaling alone drastically improves the confidence values (\eg~the expected calibration error for the I3D model drops from $15.97\%$ to $8.55\%$), our CARING method consistently leads to the best uncertainty estimates in all benchmarks, further reducing the error by $2.53\%$ on Drive\&Act.

%real driver monitoring systems, where lack of trust in such
%data-driven models remains an obstacle. This work aims at
%developing a diagnostic framework for interpreting decisions
%of such networks and can be summarized in three major
%contributions.

%Limited prior work considers ways to .... This includes t...goo

%\begin{figure}[!t]
%	\centering
%	\includegraphics[width=0.45\textwidth]{figures/density_plot.png}
%	\caption{Distributions of the Top-1 Softmax scores obtained with the I3D model for different action classes.}
%	\label{fig:density_intro}
%\end{figure}
%!TEX root = ../root.tex
\section{Related Work}

\subsection{Activity Recognition}

Activity recognition research is strongly influenced by progress in image recognition methods, where the core classification is applied on video frames and extended to deal with the video dimension on top of it. 
Similar to other computer vision fields, the methods have shifted from manually designed feature descriptors, such as Improved Dense Trajectories (IDT)~\cite{wang2013action} to Convolutional Neural Networks (CNNs) which learn intermediate representations end-to-end~\cite{wang2013action}.
The first deep learning architecture to outperform IDTs was the two-stream network~\cite{simonyan2014two, wang2016temporal}, which comprises 2D CNNs operating on individual  frames of color- and optical flow videos. 
The frame output is joined via late fusion~\cite{simonyan2014two, wang2016temporal} or an additional recurrent neural network~\cite{donahue2015long,ng2015beyond}.  
The field further progressed through emergence of 3D CNNs, which leverage spatiotemporal kernels to deal with the time dimension~\cite{ji20133d,carreira2017quo,hara2018can}. 
This type of networks still holds state-of-the-art results in the field of action recognition, with Inflated 3D Network~\cite{carreira2017quo}, 3D Residual Network~\cite{hara2018can} and Pseudo 3D ResNet~\cite{qiu2017learning} being the most prominent backbone architectures.

The above works develop algorithms with the incentive to improve the top-1 recognition accuracy on standard activity classification benchmarks \emph{without taking the faithfulness of their confidence values into account} (as demonstrated in Figure~\ref{fig:intro} with an example of Pseudo3D ResNet).
Our work focuses on \emph{uncertainty-aware action recognition} and aims for models which confidence values indeed reflect the likelihood of a correct prediction. 
Note, that the developed methods drastically improve the ability of an action recognition network to assign proper confidence values, they do not affect the accuracy, as they are based on learned scaling of the logits without changing their order.

\subsection{Identifying Model Misclassifications}

While multiple authors expressed the need for better uncertainty estimates in order to safely integrate deep CNNs in real-life systems~\cite{sunderhauf2018limits,hendrycks17baseline,nguyen2015deep},  the feasibility of predicted confidence scores has been missed out in the field of activity recognition. 
However, this problem has been addressed before in image classification~\cite{guo2017calibration,gal2016dropout}, person identification~\cite{bansal2014towards} and classical machine learning~\cite{niculescu2005predicting,degroot1983comparison,platt1999probabilistic}.
Some of the uncertainty estimation methods are handled from the Bayesian point of view, leveraging Monte Carlo Dropout sampling~\cite{gal2016dropout} or ensemble-based methods~\cite{lakshminarayanan2017simple}.
In such methods, the uncertainty is represented as a Gaussian distribution with output being the predictive mean and variance.
In contrast, calibration-based approaches~\cite{platt1999probabilistic, zadrozny2001obtaining,zadrozny2002transforming,guo2017calibration,ott2018analyzing,naeini2015obtaining} have lower computational cost as they do not preform sampling and return a single confidence value.
While these works approach the problem in a different way, they are all trained to obtain a proper confidence value on a held-out validation set following the initial training of the model and, thus, might be viewed  as postprocessing methods.
Recently, multiple calibration-based algorithms, such as isotonic regression, histogram binning, and Bayesian quantile binning regression and  were brought in the context of CNN-based image classification by Guo~\etal~\cite{guo2017calibration}.
The authors introduced temperature scaling, a simple variant of Platt Scaling~\cite{platt1999probabilistic}, where a single parameter is learned on a validation set and to rescale the neural network logits.
Despite its simplicity, the temperature scaling method has outperformed other approaches in the study by Guo~\etal~\cite{guo2017calibration} and has since then been successfully applied in natural language processing~\cite{ott2018analyzing} and medical applications~\cite{huang2020tutorial}.

Several works have studied  uncertainty  estimation  in the context of novelty detection~\cite{roitbergBMVC2018novelty,hendrycks17baseline,liang2017enhancing, roitberg2020open}.   
A Bayesian approach has been used in a framework for recognizing activity classes which were not present during training~\cite{roitbergBMVC2018novelty}. 
Hendrycks and Gimpel have introduced a baseline for  out-of-distribution detection using raw Softmax values~\cite{hendrycks17baseline}, which was further improved by Liang~\etal\cite{liang2017enhancing} through input corruptions and temperature scaling~\cite{guo2017calibration}. 
Our work, however, aims to study the confidence activity recognition models to \emph{identify, whether the prediction is correct, or not} and is therefore more comparable to the model calibration benchmarks of~\cite{guo2017calibration,niculescu2005predicting}.

Our model builds on the approach of Guo~\etal~\cite{guo2017calibration}, extending it with \emph{input-guided} scaling. 
In contrast to~\cite{guo2017calibration}, which uses a static temperature parameter for all data points, we introduce an additional \emph{calibration network} to estimate a proper scaling parameter depending on the input. 
A similar input-dependent strategy has been recently introduced in the area of pedestrian detection for autonomous driving~\cite{neumann2018relaxed}, but there are architecture-related differences to our work, \eg~, there is no additional calibration network (the scaling factor is obtained through the initial CNN) and the logits are multiplied instead of being divided by the inferred value. 
Furthermore, our benchmark examines the reliability of model confidence values in context of action recognition for the first time. 
%!TEX root = ../root.tex
\section{Uncertainty-sensitive Action Recognition}

\subsection{Problem Definition: Reliable Confidence Measures}
\label{sec:definition}
%We address the problem of supervised multi-class activity recognition with target class labels $a \in \mathcal{A}\{1,...,m\}$, where the models are usually validated only via top-1 accuracy. 
We introduce the \emph{reliability of model confidence benchmark} to supervised multi-class activity recognition, where the models are usually validated  via top-1 accuracy only~\cite{martin2019drive,kuehne2013hmdb51}. 
%Given an input video clip $x$ and a set of target class labels $a \in \mathcal{A}\{1,...,m\}$, our activity recognition model ought to not only learn to predict the correct activity (\ie~$a_{pred} = a_{true}$), but also \textit{give us well-calibrated confidence estimates} $p(a_{pred}|x)$ which indeed reflect the probability of $\mathcal{P}(a_{pred}= a_{true})$.
%Given an input video clip $x$ and a set of target class labels $a \in \mathcal{A}\{1,...,m\}$, let $m$ be our activity recognition model , with $m(x) = [a_{pred}, conf(a_{pred})]$, where $a_{pred}$ denotes the predicted activity label and $conf(a_{pred})$ the corresponding model confidence value.
Given an input video clip $x$ with a ground-truth label $a_{true}$ and the set of all possible target classes $a \in \mathcal{A}\{1,...,m\}$, let $m$ be our activity recognition model predicting an activity label $a_{pred}$ and the corresponding model confidence value $conf(a_{pred})$:  $m(x) = [a_{pred}, conf(a_{pred})]$.
A \emph{reliable} model ought to not only learn to predict the correct activity (\ie~$a_{pred} = a_{true}$), but also give us well-calibrated confidence estimates $conf(a_{pred})$, which indeed reflect the probability of a successful outcome $\mathcal{P}(a_{pred}= a_{true})$.
%learn to predict the confidence $p(a_{pred}|x)$ of the input being the activity $y_i$, which indeed reflects the probability of $a_{pred}$ being the correct prediction being $P(a_{pred}= a_{true})$.
%Given an input video clip, we wish to learn to predict the confidence $p(a_{pred}|x)$ of the input being the activity $y_i$, which indeed reflects the probability of $a_{pred}$ being the correct prediction being $P(a_{pred}= a_{true})$.
A perfectly calibrated~\ie~reliable model is often formalized as $\mathcal{P} (a_{pred}= a_{true}| conf(a_{pred}) = p) = p, \quad \forall p \in [0, 1]$ \cite{guo2017calibration}.
In other words, the inadequacy of model confidence values is directly linked to the gap between the average model confidence and model accuracy.
To quantify the calibration quality of the models' confidence scores, we use Expected Calibration Error (ECE) metric~\cite{guo2017calibration}. 
To compute ECE, we divide the space $[0, 1]$ of possible probabilities into $K$ segments (in our case, $K = 10$). 
We then compute the model accuracy and average model confidence for samples belonging to each individual segment.
In a perfectly calibrated model, the difference between accuracy and average confidence of the individual segments would be zero. 
To quantify how well we can rely on the confidence scores produced by the model, we compute the distance between the mean confidence and accuracy in each bin and then calculate the average over all such segments, weighted by the number of samples in each bin. 
Formally, the expected calibration error is defined as:
\begin{equation}
ECE = \sum_{i = 1}^{K} \frac{N_{bin_i}}{N_{total}} |acc(bin_i) - conf(bin_i)|,
\end{equation}
where $N_{bin_i}$ is the number of samples with probability values inside the bounds of $bin_i$, $acc(bin_i)$ and $ conf(bin_i)$ are the accuracy and average confidence of such examples respectively and $N_{total}$ is the total number of data points (in all bins).

\begin{figure}\centering
	\includegraphics[width=0.5\textwidth,trim={1.8cm 4.6cm 2.25cm 3.0cm},clip]{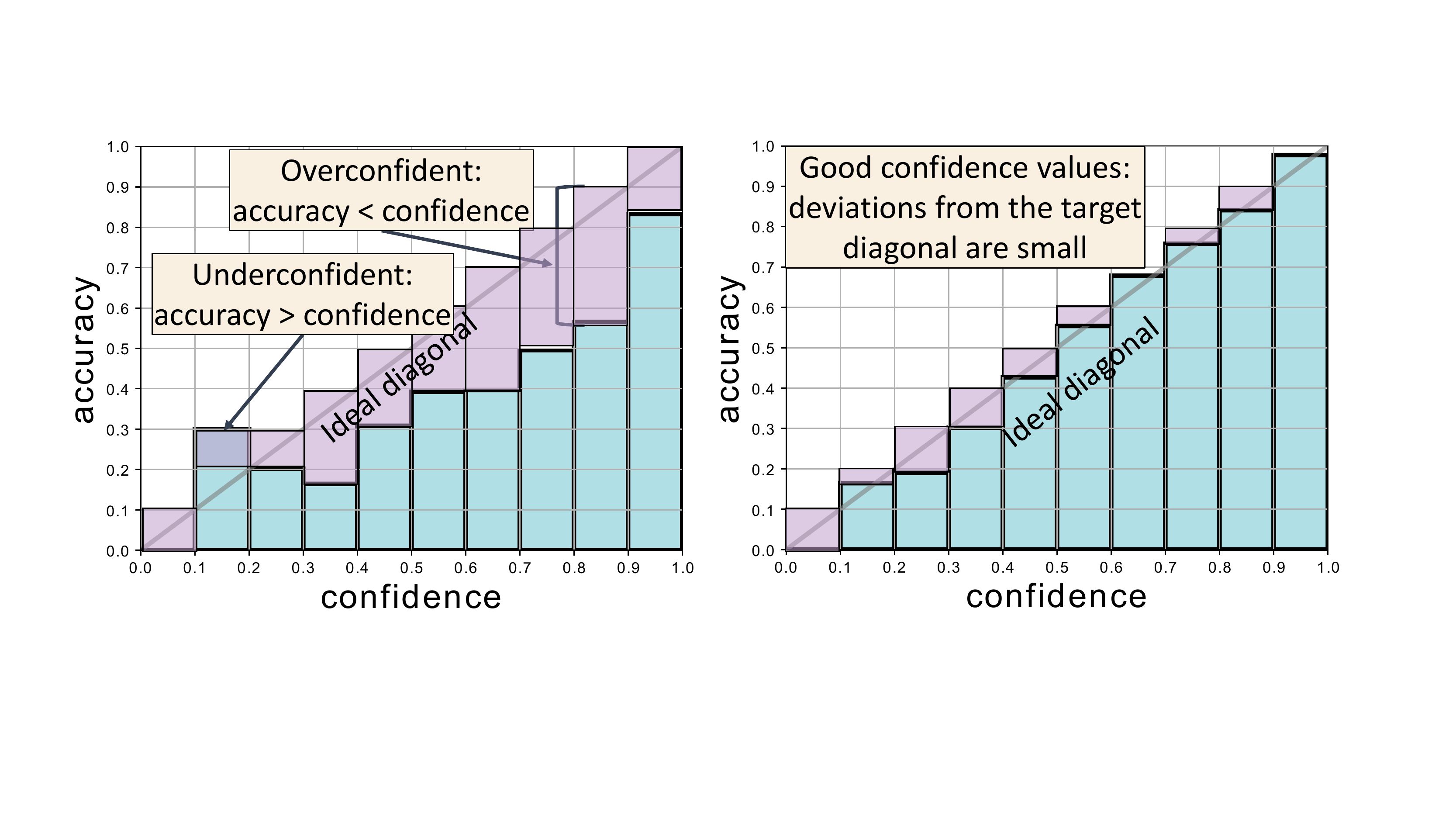}
	\caption{Reliability diagrams of a model with poor confidence estimates (left) and a well-calibrated model (right). The illustrated data are the confidence values of the Pseudo 3D ResNet a the Drive\&Act validation split before and after the improvement with the CARING calibration network.
	}
	\label{fig:reliability_explanation}
	\vspace{-0.3cm}
\end{figure}

The expected calibration error can be visualized intuitively using \emph{reliability diagrams} (example provided in Figure \ref{fig:reliability_explanation}). 
First, the space of possible probabilities (X-axis) is discretized into $K$ equally sized bins (we choose $K=10$), as previously described for the ECE calculation. 
Samples with predicted confidence  between $0$ and $0.1$ fall into the first bin, between $0.1$ and $0.2$ into the second bin and so on. 
%The X-axis represents the space of possible probabilities and is partitioned into equally sized segments, as described above. 
For each segment, we plot a bar with height corresponding to the accuracy in the current segment. 
In an ideal case, the accuracy should be equal to the average confidence score inside this bin, meaning, that the bars should have the height of the diagonal. 
As we see in Figure~\ref{fig:reliability_explanation}, these are often beyond the diagonal if the Pseudo 3D ResNet model probabilities are used out of the box. 
This means that the model tends to be overly confident, as the accuracy in the individual bins tends to be \emph{lower} than the probability produced by the model.

\subsection{Backbone Neural Architectures}
%First, we describe the backbone architectures examined in our study. 
We consider two prominent spatiotemporal CNNs for activity recognition: Inflated 3D ConvNet~\cite{carreira2017quo} and Pseudo3D ResNet~\cite{qiu2017learning}. 
%Both architectures directly operate on the video data and learn the intermediate embeddings together with the classifier layers in an end-to-end fashion.
Inflated 3D ConvNet deals with the spatial and temporal dimensions of our input by leveraging hierarchically stacked 3D-convolution and -pooling kernels with the size of  $3\times 3\times 3$.
P3D ResNet, on the other hand,  mimics 3D convolutions by applying a filter on the spatial domain ($3\times3\times1$) followed by one in the temporal dimension ($1\times1\times3$). % kernels on the (\ie~a 2D and a 1D convolution). 
%Furthermore, P3D ResNet leverages residual connections to improve the gradient flow, which allows a remarkable depth of $152$~layers, while Inflated 3D ConvNet is $27$~layers deep.
%An important characteristic of the Inflated 3D ConvNet is the use of Inception modules~\cite{szegedy2015going}, which execute convolutions with different kernel sizes simultaneously.

As in other CNNs, the neurons of the last fully-connected layer are referred to as a \emph{logit vector} $\vec{y}$ with its activations $y_a$ representing the \emph{not normalized} scores of an action $a$ being the current class.
A straight-forward way to obtain the model confidence which mimics a probability function, is to normalize the scores using Softmax:
$conf(a_{pred}) = \max\limits_{a \in \mathcal{A}} \frac{exp(y_a)}{\sum\limits_{\hat{a} \in \mathcal{A}} exp(y_{\hat{a}})}$.
During training, the cross-entropy loss is computed using the Softmax-normalized output, optimizing the network for high top-1 accuracy.
Both models have shown impressive results in activity recognition~\cite{martin2019drive,carreira2017quo,qiu2017learning}, but an evaluation of how well their Softmax-values indeed reflect the model uncertainty remains an open question and is  addressed in this work. 
%While we analyze all three models in \secref{sec:embeddings} and \secref{sec:preformance}, we choose the Inflated 3D Net for the visual explanations in \secref{sec:gradcam}, as it has shown the best recognition results in previous work.

\subsection{Calibration via Temperature Scaling}
\label{sec:temperature}

A popular way for obtaining better confidence estimates from CNN logits in image recognition is \emph{temperature scaling}~\cite{guo2017calibration}.
Temperature scaling simplifies Platt scaling~\cite{platt1999probabilistic}, and is based on learning a single parameter $\tau$ which is further used to ``soften'' the model logits.
The logits are therefore divided by $\tau$ before applying the Softmax function $\vec{y}_{scaled} = \vec{y}/\tau$.~%, where $\tau$ is a single scalar vector optimized on a held out validation set using negative-log-likelihood. 
With $\tau>1$ the resulting probabilities become smoother, moving towards $\frac{1}{m}$, where $m$ is the number of classes. 
Contrary, scaled probability would approach $1$ as $\tau$ becomes closer to $0$.
After the neural network is trained for supervised classification in a normal way, we fix the model weights and optimize $\tau$ on a held-out validation set using Negative-Log-Likelihood. 
Despite method simplicity, temperature scaling has been  highly effective for obtaining well-calibrated image recognition CNNs, surpassing heavier methods such as Histogram binning and Isotonic Regression~\cite{guo2017calibration}.
%Still, its potential for spatiotemporal video classification CNNs has not been explored yet.

As this method has not been explored  for spatiotemporal video classification CNNs yet, we augment the Inflated 3D ConvNet and Pseudo 3D ResNet models with a post-processing temperature scaling module. 
We optimize $\tau$ using Gradient Descent with a learning rate of $0.01$ for $50$ epochs.
 
Note, that as the networks are fully trained and their weights remain fixed while learning the scaling parameter $\tau$, transformation of the logits does not influence their order and therefore the \emph{model accuracy stays the same}. 
In other words, while temperature scaling gives us better uncertainty estimates, the predicted activity class does not change as all logits are divided by the same scalar. 
%post-processing

%\begin{equation}
%p(a_{pred}) = \max\limits_{a \in \mathcal{A}} \frac{exp(y_a/T)}{\sum\limits_{\hat{a} \in \mathcal{A}} exp(y_{\hat{a}}/T)}
%\end{equation}

\subsection{Calibrated Action Recognition with Input Guidance}
\label{sec:caring}
\begin{figure}[!t]
	\centering
	\includegraphics[trim={0 5.8cm 6.5cm 0},clip,width=0.5\textwidth]{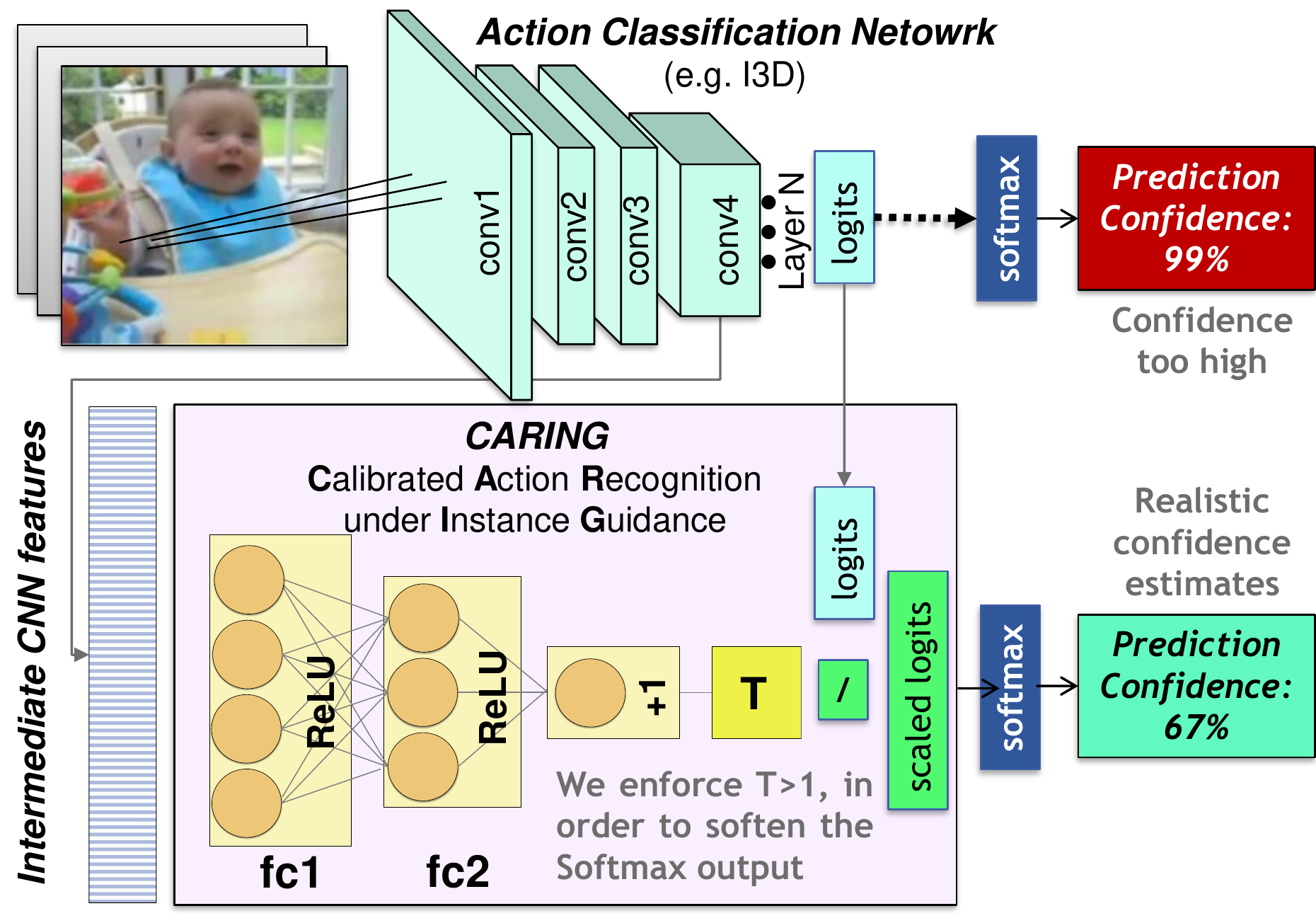}
	\caption{Overview of the Calibrated Action Recognition under Instance Guidance Model (CARING). CARING is an additional neural network which learns to infer the scaling factor $\mathcal{T}$ depending on the instance representation. The logits of the original activity recognition network are then divided by $T$, giving better estimates of the model uncertainty. }
	\label{fig:caring-model}
		\vspace{-0.3cm}
\end{figure}

%Figure \ref{fig:learning_procedure} illustrates the evolution of the Expected Calibration Error (defined in Section \ref{sec:definition}) and the resulting average and standard deviation of the inferred scaling parameter  $\mathcal{T}(\vec(z))$ in the validation data.
%Figure \ref{fig:avg_temperature_during_training} reveals, that both, the mean and standard deviation of temperature rises during training, leading to an improvement in the calibration error (see Figure \ref{fig:ece_during_training}). 
%The observed increase in the standard deviation of the scaling parameter confirms that handling the logits differently dependent on the input is beneficial in our task.

In this section, we introduce a new model for obtaining proper confidence estimates by learning how to scale the logits \emph{depending on the input}.
%While our evaluation described in the next section reveals, that previous method clearly improves model confidence calibration, it does not take into account representation of the current example when deciding, how to scale the output and the logits are always divided by the same global scalar $\tau$.
While our evaluation described in the next section reveals, that previous method clearly improves model confidence calibration, it does not take into account representation of the current example, \ie,  the logits are always divided by the \textit{same} global scalar $\tau$.

We believe, that the input itself carries useful signal for inferring model confidence and build on the temperature scaling approach~\cite{guo2017calibration} with one crucial difference: the scaling factor is not global but different for varying input.   
Our main idea is therefore to learn acquiring the scaling parameter $\mathcal{T}(\vec{z})$ on-the-fly at test-time depending on the input representation $\vec{z}$, so that the scaled logits become $\vec{y}_{scaled} = \vec{y}/\mathcal{T}(\vec{z})$. 
To learn the input-dependent temperature value $\mathcal{T}(\vec{z})$, we introduce an additional \emph{calibration neural network}, which we refer to as the CARING model (\textbf{C}alibrated \textbf{A}ction \textbf{R}ecognition under \textbf{In}put \textbf{G}uidance), as it guides the scaling of the logits depending on the current instance.
An overview of our model is provided in Figure \ref{fig:caring-model}.
CARING network comprises two fully-connected layers, with  the output of the second layer being a single neuron used to infer the input-dependent temperature scalar.    
Note, that we extend the last ReLU activation with an addition of $1$ to enforce $\mathcal{T}(\vec{z})\geq1$, required to soften the probability scores.
Input-dependent temperature $\mathcal{T}(\vec{z})$ is therefore obtained as:
\begin{equation}
\quad \mathcal{T}(\vec{z}) = 1 +relu( W_2~relu( W_1 \vec{z}+\vec{b}_1)  + \vec{b_2}), 
\end{equation}
 where $W_1$,$W_2$, $b_1$ and $b_2$ are the network weight matrices and bias vectors and $\vec{z}$ is the input representation, for which we use the intermediate features of the original activity recognition network ($\vec{z}$ has a size of $1024$ for Infalted 3D ConvNet and $2048$ for Pseudo 3D ResNet).
 We then scale the logits by the inferred instance-dependent temperature $\mathcal{T}(\vec{z})$ and our prediction probability becomes:
%\begin{equation}
%p(a_{pred}) = \max\limits_{a \in \mathcal{A}} \frac{exp(y_a/\mathcal{T}(z))}{\sum\limits_{\hat{a} \in \mathcal{A}} exp(y_{\hat{a}}/\mathcal{T}(z))}
%\end{equation}
\begin{equation}
conf(a_{pred}) = \max\limits_{a \in \mathcal{A}} \frac{exp(\frac{y_a}{\mathcal{T}(\vec{z})})}{\sum\limits_{\hat{a} \in \mathcal{A}} exp(\frac{y_{\hat{a}}}{\mathcal{T}(\vec{z})})}.
\end{equation}

We train the CARING model on a held-out validation set with Negative Log Likelihood loss for $300$ epochs (learning rate of $0.005$, weight decay of $1e^{-6}$).
Although ${\mathcal{T}(\vec{z})}$ is not a constant and varies depending on the input, the order of the output neurons stays the same, since the CARING model infers one single value given an input $\vec{z}$, so that all logits are divided by the same value ${\mathcal{T}(\vec{z})}$. 
Similarly to the approach described in Section \ref{sec:caring}, CARING can be viewed as a post-processing step for obtaining better uncertainty confidence and \emph{does not affect the predicted activity class and model accuracy}. 

\begin{figure}
	\centering
	\begin{subfigure}{.25\textwidth}
		\centering
		\includegraphics[trim={0.4cm 0 0 0},clip,width=\linewidth]{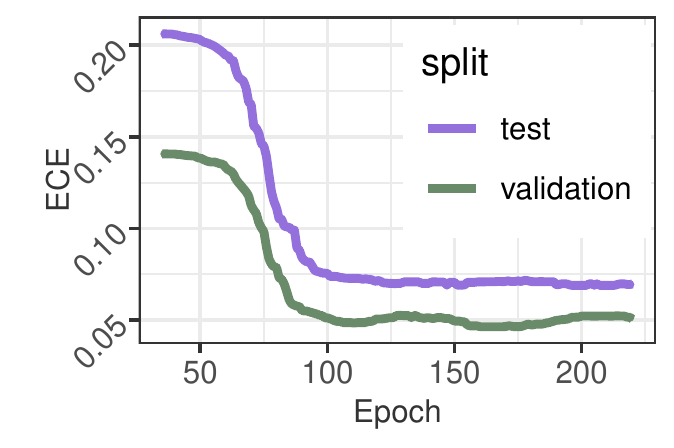}
		\caption{Expected Calibration Error improvement during the training procedure for validation  and test data.}
		\label{fig:ece_during_training}
	\end{subfigure}%
	\begin{subfigure}{.25\textwidth}
		\centering
		\includegraphics[trim={0.4cm 0 0 0},clip,width=\linewidth]{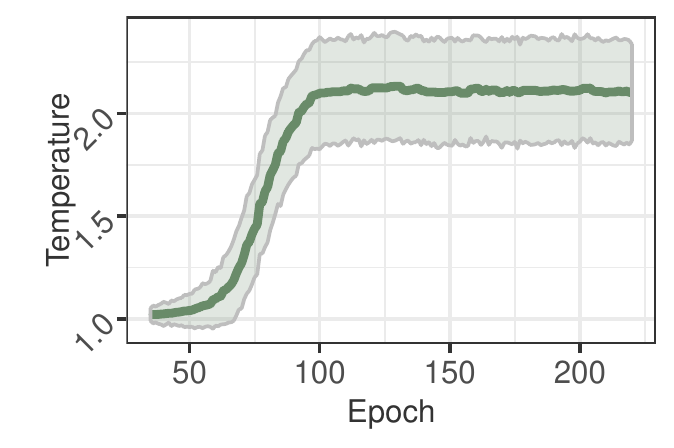}
		
		\caption{Average temperature and its\\ \null \quad standard deviation estimated by \\ \null \quad our model during training.}
		\label{fig:avg_temperature_during_training}
	\end{subfigure}
	\caption{CARING model evolution during training for one Drive\&Act split. Both average value and standard deviation of the learned input-dependent scaling parameter  $\mathcal{T}\vec(z)$ rise as the training proceeds (right figure). Jointly with the decrease of the calibration error (left figure), this indicates the usefulness of learning different scaling parameters for different inputs.}
	\label{fig:learning_procedure}
		\vspace{-0.3cm}
	
\end{figure}

We validate, that learning input-dependent temperature value is indeed better than using a single global scaling parameter by examining the evolution of different model metrics during training.
Figure~\ref{fig:learning_procedure} illustrates changes of the expected calibration error (defined in Section \ref{sec:definition}) and the  average and standard deviation of the inferred scaling parameter  $\mathcal{T}\vec(z)$ measured over the validation data.
%Figure \ref{fig:learning_procedure} illustrates a of the expected calibration error (defined in Section \ref{sec:definition}) and the resulting average and standard deviation of the inferred scaling parameter  $\mathcal{T}(\vec(z))$ in the validation data.
Figure~\ref{fig:avg_temperature_during_training} reveals, that both, the mean and standard deviation of temperature rises during training, leading to a lower calibration error (Figure~\ref{fig:ece_during_training}). 
The observed increase in the standard deviation of the scaling parameter confirms that handling the logits differently dependent on the input is beneficial in our task.
%!TEX root = ../root.tex

\section{Experiments}

\subsection{Benchmark settings}
\label{sec:benchmark-settings}

Since  there  is  no  established  evaluation  procedure  targeting the reliability of confidence values in context of activity recognition, we adapt existing evaluation protocols for two conventional action classification datasets, Drive\&Act~\cite{martin2019drive} and HMDB-51~\cite{kuehne2013hmdb51}, for our task.
%We analyze confidence score reliability of activity recognition methods using two prominent datasets. 
%First, we use the Drive\&Act~\cite{martin2019drive} benchmark for driver activity recognition, as its application-driven challenges, such as fine-grained categories and unbalanced data distribution are typical for real-life systems.
We choose the Drive\&Act~\cite{martin2019drive} testbed for driver activity recognition as our main benchmark, as it is application-driven and encompasses multiple challenges typical for real-life systems (\eg~fine-grained categories and unbalanced data distribution).
Drive\&Act comprises $34$ fine-grained activity classes, which, however are highly unbalanced as the number of examples ranged from only $19$ examples of \emph{taking laptop from backpack} to $2797$  instances of \emph{sitting still}.
As CNNs have a lower performance when learning from few examples, we sort the behaviors by their frequency in the dataset and divide them into \emph{common} (top half of the classes) and rare (the bottom half).
We subsequently evaluate the models in three modes: considering \emph{all activities}, as it is usually done, using only the \emph{overrepresented-} or only the \emph{rare} classes. \looseness=-1
%!TEX root = ../root.tex

%\begin{table}[!htbp]
\begin{table}[!t]

	\scalebox{1.02}{
	\begin{tabular}{ @{}lcccc}
		\toprule
		\multirow{2}{*}{\textbf{Model}} & \multicolumn{2}{c}{\textbf{ECE}} & \multicolumn{2}{c}{\textbf{NLL}} \\
		& val & test & val & test  \\ 
		\midrule
		\textbf{Drive\&Act - Common Classes} \\
		P3D \cite{qiu2017learning} \scriptsize{\textcolor{table_standard}{\textbf{\circled{S}}}}&16.9 & 19.39 & 1.63 & 1.85  \\
		I3D \cite{carreira2017quo} \scriptsize{\textcolor{table_standard}{\textbf{\circled{S}}}}&10.22 & 13.38 & 0.90 & 1.27 \\
		%C3D & X & X & X & X & X & X  \\
		P3D + Temperature Scaling  \cite{guo2017calibration}\scriptsize{\textcolor{table_uncertainty}{\textbf{\circled{U}}}}&5.65 & 5.7 & 1.28 & 1.48\\
		I3D + Temperature Scaling \cite{guo2017calibration}\scriptsize{\textcolor{table_uncertainty}{\textbf{\circled{U}}}}&5.31 & 6.99 & 0.57 & 0.83  \\
		CARING - P3D (ours) \scriptsize{\textcolor{table_uncertainty}{\textbf{\circled{U}}}} &4.81 & \textbf{4.27} & 1.19 & 1.42\\
		CARING - I3D (ours) \scriptsize{\textcolor{table_uncertainty}{\textbf{\circled{U}}}}&\textbf{2.57 }& 5.26 & \textbf{0.50} & \textbf{0.78}  \\
		\midrule
		\textbf{Drive\&Act - Rare Classes} &  &  & &  \\
		P3D \cite{qiu2017learning} \scriptsize{\textcolor{table_standard}{\textbf{\circled{S}}}}&31.49 & 37.25 & 3.43 & 4.68 \\
		I3D \cite{carreira2017quo} \scriptsize{\textcolor{table_standard}{\textbf{\circled{S}}}}&31.48 & 43.32 & 3.41 & 4.54\\
		%C3D & X & X & X & X & X & X  \\
		P3D + Temperature Scaling \cite{guo2017calibration} \scriptsize{\textcolor{table_uncertainty}{\textbf{\circled{U}}}}&17.83 & 21.09 & 2.26 & 2.99\\
		I3D + Temperature Scaling \cite{guo2017calibration} \scriptsize{\textcolor{table_uncertainty}{\textbf{\circled{U}}}}&24.97 & 32.38 & 1.96 & 2.62\\
		CARING - P3D  (ours) \scriptsize{\textcolor{table_uncertainty}{\textbf{\circled{U}}}}&\textbf{13.73 }& \textbf{19.92} & 2.12 & 2.93\\
		CARING - I3D (ours) \scriptsize{\textcolor{table_uncertainty}{\textbf{\circled{U}}}}&18.34 & 23.6 & \textbf{1.55} & \textbf{2.17}\\
		\midrule
		\textbf{Drive\&Act - All Classes} \\
		P3D \cite{qiu2017learning} \scriptsize{\textcolor{table_standard}{\textbf{\circled{S}}}}&17.89 & 21.09 & 1.77 & 2.12 \\
		I3D \cite{carreira2017quo} \scriptsize{\textcolor{table_standard}{\textbf{\circled{S}}}}&11.72 & 15.97 & 1.10 & 1.56 \\
		%C3D & X & X & X & X & X & X  \\
		P3D + Temperature Scaling  \cite{guo2017calibration} \scriptsize{\textcolor{table_uncertainty}{\textbf{\circled{U}}}}&5.89 & 6.41 & 1.35 & 1.63\\
		I3D + Temperature Scaling \cite{guo2017calibration} \scriptsize{\textcolor{table_uncertainty}{\textbf{\circled{U}}}}&6.59 & 8.55 & 0.68 & 0.99\\
		CARING - P3D (ours) \scriptsize{\textcolor{table_uncertainty}{\textbf{\circled{U}}}}&4.58 & \textbf{5.26} & 1.26 & 1.57 \\
		CARING - I3D (ours) \scriptsize{\textcolor{table_uncertainty}{\textbf{\circled{U}}}}&\textbf{3.03} & 6.02 & \textbf{0.58} & \textbf{0.9} \\ 
		\midrule
		\textbf{HMDB-51} \\
		I3D \cite{carreira2017quo} \scriptsize{\textcolor{table_standard}{\textbf{\circled{S}}}}& 10.29 & 20.11 & 0.98&1.97 \\
		I3D + Temperature Scaling \cite{guo2017calibration} 	\scriptsize{\textcolor{table_uncertainty}{\textbf{\circled{U}}}} &	4.00& 7.75 &\textbf{0.81 }&1.57 \\
		CARING - I3D (ours) \scriptsize{\textcolor{table_uncertainty}{\textbf{\circled{U}}}} & \textbf{3.38} & \textbf{5.98} & \textbf{0.81} & \textbf{1.54} \\ 
		\multicolumn{5}{c}{\scriptsize{\textcolor{table_standard}{\textbf{\circled{S}}}} Standard activity recognition models\quad \quad \quad \quad  \scriptsize{\textcolor{table_uncertainty}{\textbf{\circled{U}}}}  Uncertainty-aware models}
		\\
		\bottomrule

	\end{tabular}
		}
		\caption{Reliability of confidence values on the Drive\&Act~\cite{martin2019drive} and HMDB-51~\cite{kuehne2013hmdb51} datasets for standard activity recognition models and their extensions with uncertainty-aware calibration algorithms.
	}
	\vspace{-0.3cm}
	\label{tbl:reliability_results}

\end{table}

We further validate the models on  HMDB-51~\cite{kuehne2013hmdb51}, a more general activity recognition dataset comprising of YouTube videos.
The benchmark covers $51$ activity classes, which are more discriminative in their nature (\eg~laughing and playing football) and are perfectly balanced (three splits with $70$ training and $30$ test examples for every category).

Input to the P3D- and I3D models are snippets of $64$ consecutive frames. 
If the original video segment is longer, the snippet is chosen randomly during training and at the video center at test-time. 
If the video segment is shorter, we repeat the last frame until the $64$ frame snippet is filled.

Following the problem definition of Section~\ref{sec:definition}, we extend the standard accuracy-driven evaluation protocols~\cite{martin2019drive,kuehne2013hmdb51} with the expected calibration error (ECE), depicting the deviation of model confidence score from the true misclassification probability.
In addition, we report the Negative Log Likelihood (NLL), as high NLL values are linked to model miscalibration~\cite{guo2017calibration}.
%As done in the original work~\cite{martin2019drive}, we use the three Drive\&Act splits with distinct people for training (10~people), test (3~people) and validation (2~people). 
Since HMDB-51 does not contain a validation split, we randomly separate $10$\% of the training data for this purpose. 
As done in the original works~\cite{martin2019drive, kuehne2013hmdb51}, we report the average results over the three splits for both testbeds.

%Previous evaluation of CNN-based models inside the vehicle cabin has focused on the multi-class top-1 accuracy as a single performance metric~\cite{MartinRoitberg2019}.
%This is an oversimplification as the prediction quality varies greatly depending on multiple factors, that we are going to uncover in this section.
%To examine the strengths and weaknesses of CNN-based algorithms, we extend the evaluation procedure of~\cite{MartinRoitberg2019} with multiple settings and metrics.

%%%%%%%%%%

\subsection{Confidence Estimates for Action Recognition}
\label{sec:exp-closed-set}
%!TEX root = ../root.tex

\begin{table*}[!t]
%\begin{table*}[!htbp]
	\centering
	%\vspace{0.1cm}	

	\scalebox{1.23}{	
\begin{tabular}{lllcccccc}
	\toprule
	\multirow{2}{*}{Activity}       & \multirow{2}{*}{\begin{tabular}[c]{@{}l@{}}Number of\\ Samples\end{tabular}} & \multirow{2}{*}{Recall} & \multicolumn{3}{c}{I3D \scriptsize{\textcolor{table_standard}{\textbf{\circled{S}}}}}      & \multicolumn{3}{c}{CARING-I3D \scriptsize{\textcolor{table_uncertainty}{\textbf{\circled{U}}}}} \\
	&                                                                              &                         & Mean Conf. & $\Delta$Acc & ECE   & Mean Conf.  & $\Delta$Acc  & ECE   \\
	\midrule
	\multicolumn{2}{l}{\textbf{Five most common activities}}    &&&&&&&                                                                                                                                                       \\
	sitting\_still                  & 2797                                                                         & 95.1                    & 97.96     & 2.86     & 2.86  & 93.84      & -1.26     & \textbf{1.84}\\
	
	eating                          & 877                                                                          & 86.42                   & 93.26     & 6.84     & 9.33  & 80.99      & -5.43     & \textbf{5.75} \\
	fetching\_an\_object            & 756                                                                          & 76.03                   & 93.77     & 17.74    & 18.28 & 79.42      & 3.4       &\textbf{5.32}  \\
	%interacting\_with\_phone        & 471                                                                          & 85.49                   & 96.04     & 10.55    & 10.55 & 84.67      & -0.82     & 3.28  \\
	placing\_an\_object             & 688                                                                          & 66.77                   & 93.03     & 26.25    & 26.25 & 75.9       & 9.13      & \textbf{9.25}  \\
	reading\_magazine               & 661                                                                          & 92.93                   & 98.58     & 5.65     & 6.09  & 93.35      & 0.42      & \textbf{2.87}  \\
	%reading\_newspaper              &                                                                              & 94.47                   & 97.32     & 2.85     & 4.19  & 87.68      & -6.79     & 7.4   \\

	%talking\_on\_phone              & 359                                                                          & 71.6                    & 92.55     & 20.95    & 22.96 & 78.26      & 6.65      & 13.36 \\
	%using\_multimedia\_display      & 474                                                                          & 92.39                   & 97.38     & 4.99     & 4.99  & 91.25      & -1.14     & 2.28  \\
	%working\_on\_laptop             & 392                                                                          & 86.86                   & 94.33     & 7.48     & 9.51  & 85.67      & -1.19     & 3.89  \\
	\midrule
	\multicolumn{2}{l}{\textbf{Five most underrepresented activities}}   &&&&&&&                                                                                                                                                   \\
	closing\_door\_inside           & 30                                                                           & 92.31                   & 98.51     & 6.21     & \textbf{8.22}  & 86.00         & -6.31     & 8.30   \\
	closing\_door\_outside          & 22                                                                           & 81.82                   & 93.55     & 11.73    & 20.97 & 86.86      & 5.04      & \textbf{19.81} \\
	%closing\_laptop                 & 40                                                                           & 53.85                   & 87.52     & 33.67    & 35.96 & 65.41      & 11.57     & 21.71 \\
	%entering\_car                   & 41                                                                           & 100                     & 99.25     & -0.75    & 0.75  & 91.31      & -8.69     & 8.69  \\
	%exiting\_car                    & 43                                                                           & 100                     & 95.17     & -4.83    & 4.83  & 91.19      & -8.81     & 8.81  \\
	opening\_backpack               & 27                                                                           & 0                       & 98.82     & 98.82    & 98.82 & 82.69      & 82.69     & \textbf{82.69} \\
	%opening\_door\_inside           & 39                                                                           & 81.25                   & 93.51     & 12.26    & 12.26 & 81.18      & -0.07     & 10.2  \\
	%opening\_door\_outside          & 30                                                                           & 100                     & 100       & 0        & 0     & 98.16      & -     & 1.84  \\
	putting\_laptop\_into\_backpack & 26                                                                           & 16.67                   & 92.67     & 76.00       & 76.00    & 76.46      & 59.8      & \textbf{59.80}  \\
	taking\_laptop\_from\_backpack  & 19                                                                           & 0.00                       & 85.25     & 85.25    & 85.25 & 70.08      & 70.08     & \textbf{70.08}\\
	\multicolumn{9}{c}{\scriptsize{\textcolor{table_standard}{\textbf{\circled{S}}}} Standard activity recognition models\quad \quad \quad \quad  \scriptsize{\textcolor{table_uncertainty}{\textbf{\circled{U}}}}  Uncertainty-aware models}\\
	\bottomrule
\end{tabular}
	}
	\caption{
	Analysis of the resulting confidence estimates of the initial I3D model and its CARING version for individual common and rare Drive\&Act activities. \emph{Recall} denotes the recognition accuracy of the current class, while \emph{Mean Conf.} denotes the average confidence estimate produced by the model. Supplemental to the Expected Calibration Error (\emph{ECE}), we report the difference between the mean confidence value and model accuracy (denoted \emph{$\Delta$Acc}). While in a perfectly calibrated model $\Delta$Acc is 0, ECE is a better evaluation metric, as \eg~if a lot of samples have too high and too low confidence values, their average might lead to a misconception of good calibration. While there is room for improvement for underrepresented and poorly recognized activity classes, the CARING model consistently leads to better uncertainty estimates. 
}
	
	\label{tbl:class-wise}
	%\vspace{-0.3cm}
	
\end{table*}
%In Table \ref{tbl:reliability_results} we report the performance of the CNN-based Inflated 3D ConvNet (I3D) and Pseudo 3D ResNet (P3D) approaches and their uncertainty-aware versions in terms of the ECE and NLL for rare, overrepresented and all $Drive\&Act$ activity classes as well as different I3D-based models on the HMDB-51 dataset.
In Table \ref{tbl:reliability_results} we compare CNN-based activity recognition approaches and their uncertainty-aware versions in terms of the expected calibration error and NLL for \textit{rare}, \textit{overrepresented} and \textit{all} Drive\&Act classes as well as in the HMDB-51 setup.
First, we verify our suspicion that native activity recognition architectures provide unreliable confidence estimates: confidence scores produced by I3D score have a misalignment of $15.97\%$ for Drive\&Act  and $20.11\%$ for HMDB-51. 
Similar issues are present in P3D: $21.2\%$ ECE on Drive\&Act, an error far too high for safety-critical applications.

%\vspace{-0.4cm}

Model reliability is clearly improved by learning to obtain proper probability estimates, as all uncertainty-aware variants surpass the raw Softmax values.
%Model reliability is improved by using uncertainty-aware models, \ie learning to obtain proper probability estimates by temperature scaling or our CARING model (see Sections and \ref{sec:temperature} \ref{sec:caring}).
Interestingly, although I3D  has better initial uncertainty estimates than P3D (ECE of $21.09$\% for P3D, $15.97$\% for I3D), P3D seems to have a stronger response to both, temperature scaling  and CARING approaches then I3D (ECE of $5.26$\% for CARING-P3D, $6.02$\% for CARING-I3D). 
However, as this difference is very small ($<1\%$), we would rather recommend using I3D, as it mostly gives higher accuracy~\cite{martin2019drive,carreira2017quo,qiu2017learning}.
While we consider the expected calibration error  to be of vital importance for applications, we realize that this metric is complementary to model accuracy and encourage taking both measures into account when selecting the right model.
We want to remind, that both temperature scaling and the CARING method \emph{do not influence the model accuracy} (see Sections \ref{sec:temperature} and \ref{sec:caring}). 
%I3D test = 63.09 63.64 0.55
%I3D Val = 68.71 69.57 0.86
%P3D Val = 54.86 55.04 0.18
%P3D Test = 46.62 45.32 1.3%
For Pseudo 3D ResNet we achieve an overall accuracy of $54.86\%$ (validation) and $46.62\%$ (test) on Drive\&Act, which does not change through our uncertainty-based modifications.
%Consistently with previous work~\cite{martin2019drive} we observe a clearly higher recognition rate of I3D and its variants with $68.71\%$ (validation) and $63.09\%$ (test) accuracy\footnote{The slight deviation from the accuracy reported in the original work\cite{martin2019drive} (between $0.18\%$ and $1.3\%$) is due to random factors in the training process.}.
Consistently with~\cite{martin2019drive} I3D achieves a higher accuracy of $68.71\%$ for validation and $63.09\%$ for test set \footnote{The slight deviation from the accuracy reported in the original work\cite{martin2019drive} (between $0.18\%$ and $1.3\%$) is due to random factors in the training process.}.

As expected, the model confidence reliability correlates with the amount of training data (see distinguished areas for \textit{common}, \textit{rare} and \textit{all} classes of Drive\&Act in Table \ref{tbl:reliability_results}). 
For example, the \emph{common classes} setting encounters the lowest expected calibration error for both original and uncertainty-aware architectures ($13.38\%$ for I3D, $5.26\%$ for CARING-I3D).
Leveraging intermediate input representation via our CARING calibration network leads to the best probability estimates on both datasets and in all evaluation settings. 
Thereby, the CARING strategy surpasses the raw neural network confidence by $9.95\%$ and the temperature scaling method by $2.53\%$ on Drive\&Act, highlighting the usefulness of learning to obtain probability scores \emph{depending on the input}.

%While the Inflated 3D ConvNet  outperforms other approaches in all metrics ($63.64\%$  top-1 test accuracy for all classes), C3D seems to be stronger than Pseudo 3D ResNet in terms of the top-5 accuracy, while the latter model is better in top-1 classification. 
%C3D therefore is well suited for coarse classification but has issues discovering fine-grained structures.  
%While it is expected, that the top-1 recognition rate is significantly lower than the top-5 results, this gap grows by a large margin for rare classes (\eg this difference is $32.52\%$ for uncommon- and $17.18\%$ for common actions when considering the Inflated 3D ConvNet test setting).
We further examine model performance for the individual
classes, considering the five most frequent and the five most uncommon Drive\&Act activities separately in Table \ref{tbl:class-wise}.
 In addition to ECE, we report the accuracy for samples belonging to the individual class, the average confidence value they obtained with the corresponding model and the difference between them (denoted \emph{$\Delta$Acc}). 
 While a such global confidence-accuracy disagreement is interesting to consider (and is  $0$ for a perfectly calibrated model) it should be viewed with caution, as it might lead to an incorrect illusion of good confidence calibration, as \eg a lot of samples with too high and too low confidence values might cancel each other out through averaging. \looseness=-1
 
Reliability of the confidence scores is significantly improved through the CARING method and is connected to the amount of training data and the accuracy. 
Models have significant issues with learning from few examples (\eg~$76\%$ I3D and $59.80\%$ CARING-I3D ECE for \emph{putting laptop into backpack}). 
 For both, over- and underrepresented classes, the ECE of easy-to-recognize activities (\ie the ones with high accuracy) is lower. 
 %One interesting observation is that before calibration, average confidence score is higher than the accuracy for all classes (positive $\Delta Acc$), while after logit transformation with the CARING model the average confidence score is lower than the accuracy for some classes, such as \emph{eating}. 
Before calibration, the average confidence value is always higher than the accuracy (positive $\Delta Acc$) disclosing that the models are too optimistic in their predictions.
Interestingly, after the CARING transformation is applied, the average model  confidence is lower than the accuracy for some classes, such as \emph{eating}. 
CARING models therefore tend to be more conservative in their assessment of certainty.

\subsection{Calibration Diagrams}
\label{sec:calib_diag}

\begin{figure*}[]
	\centering
	\begin{subfigure}{.16\textwidth}
		\centering
		\includegraphics[trim={0.1cm 0 1.4cm 0},clip,width=\linewidth]{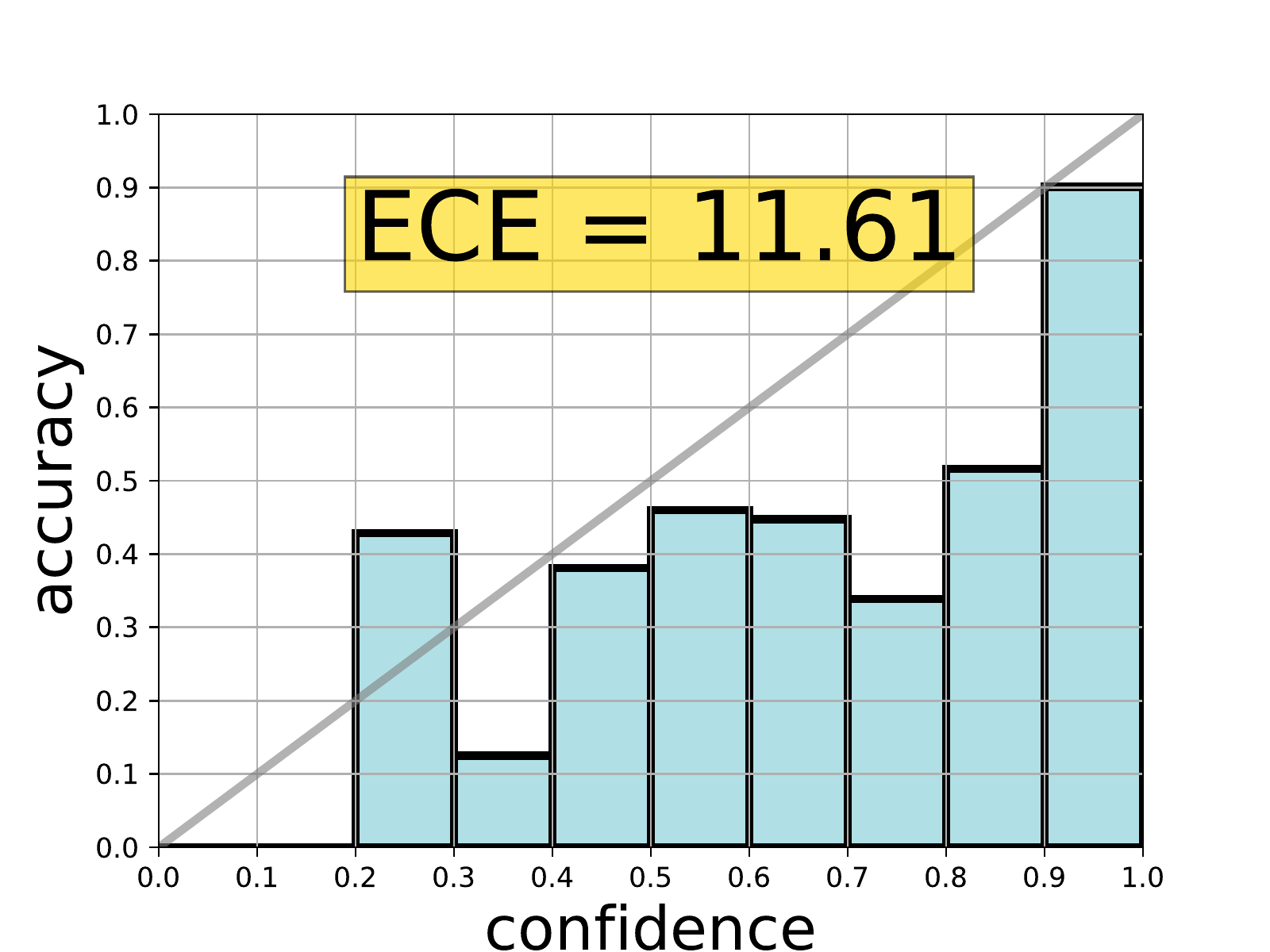}
		\caption{I3D (original), \\ \null \quad all  action classes}
		\label{fig:i3d_all}
	\end{subfigure}%
	\begin{subfigure}{.16\textwidth}
		\centering
		\includegraphics[trim={0.1cm 0 1.4cm 0},clip,width=\linewidth]{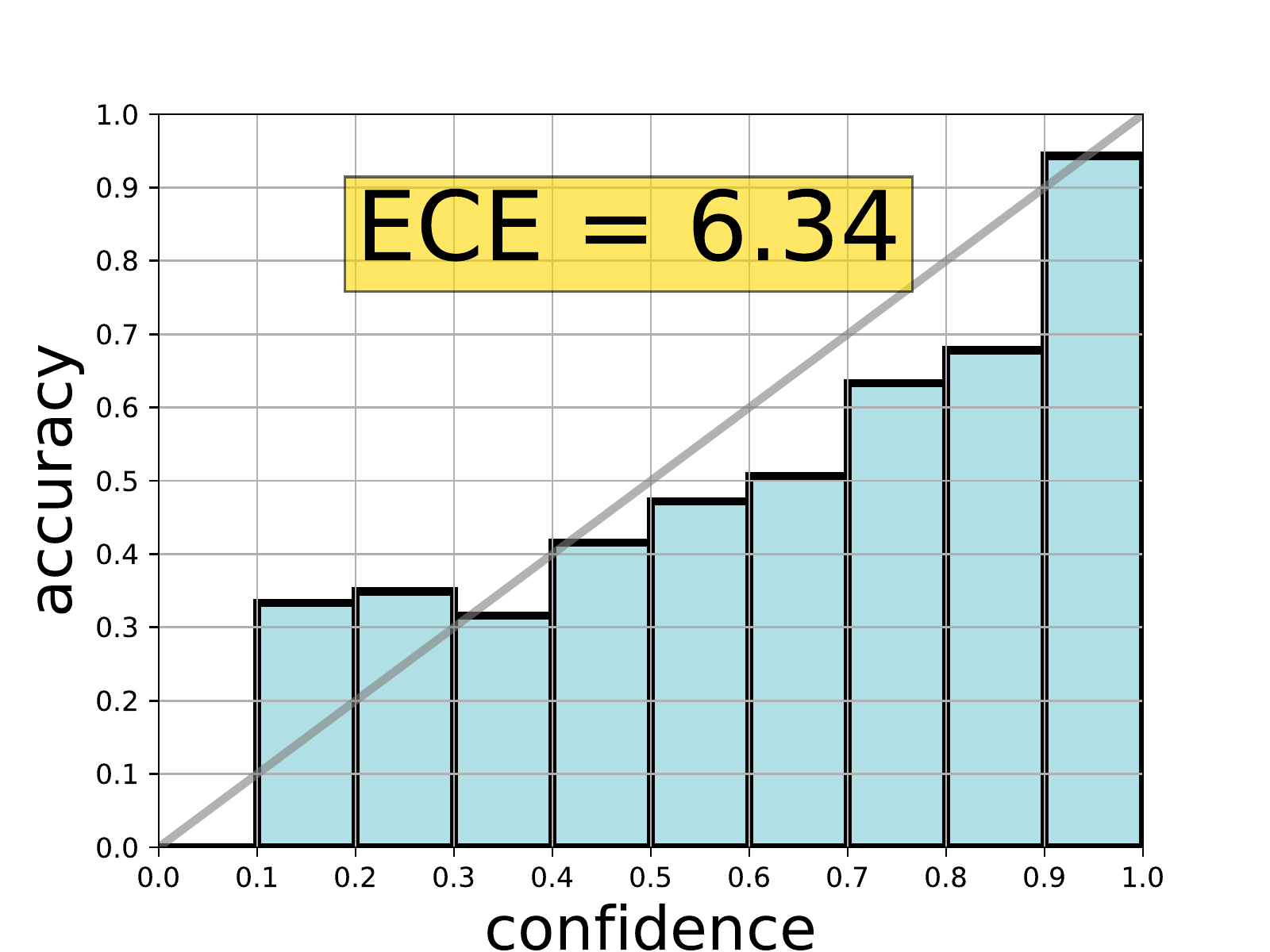}
		\caption{I3D + temp. scaling, all action classes}
		\label{fig:i3d_temp_all}
	\end{subfigure}%
	\begin{subfigure}{.16\textwidth}
		\centering
		\includegraphics[trim={0.1cm 0 1.4cm 0},clip,width=\linewidth]{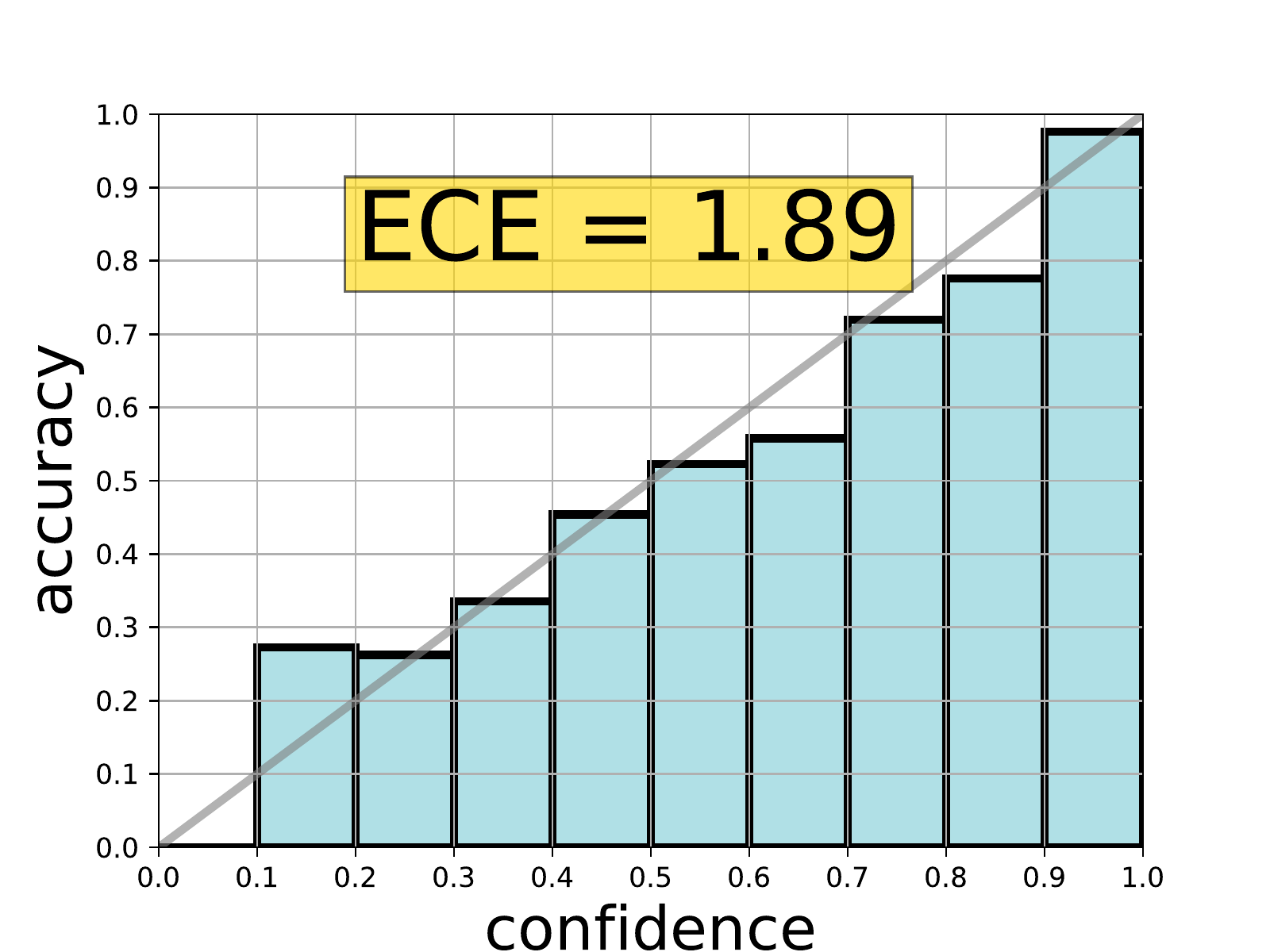}
		\caption{CARING-I3D, \\ \null \quad all action classes}
		\label{fig:i3d_caring_all}
	\end{subfigure}%
	\begin{subfigure}{.16\textwidth}
		\centering
		\includegraphics[trim={0.1cm 0 1.4cm 0},clip,width=\linewidth]{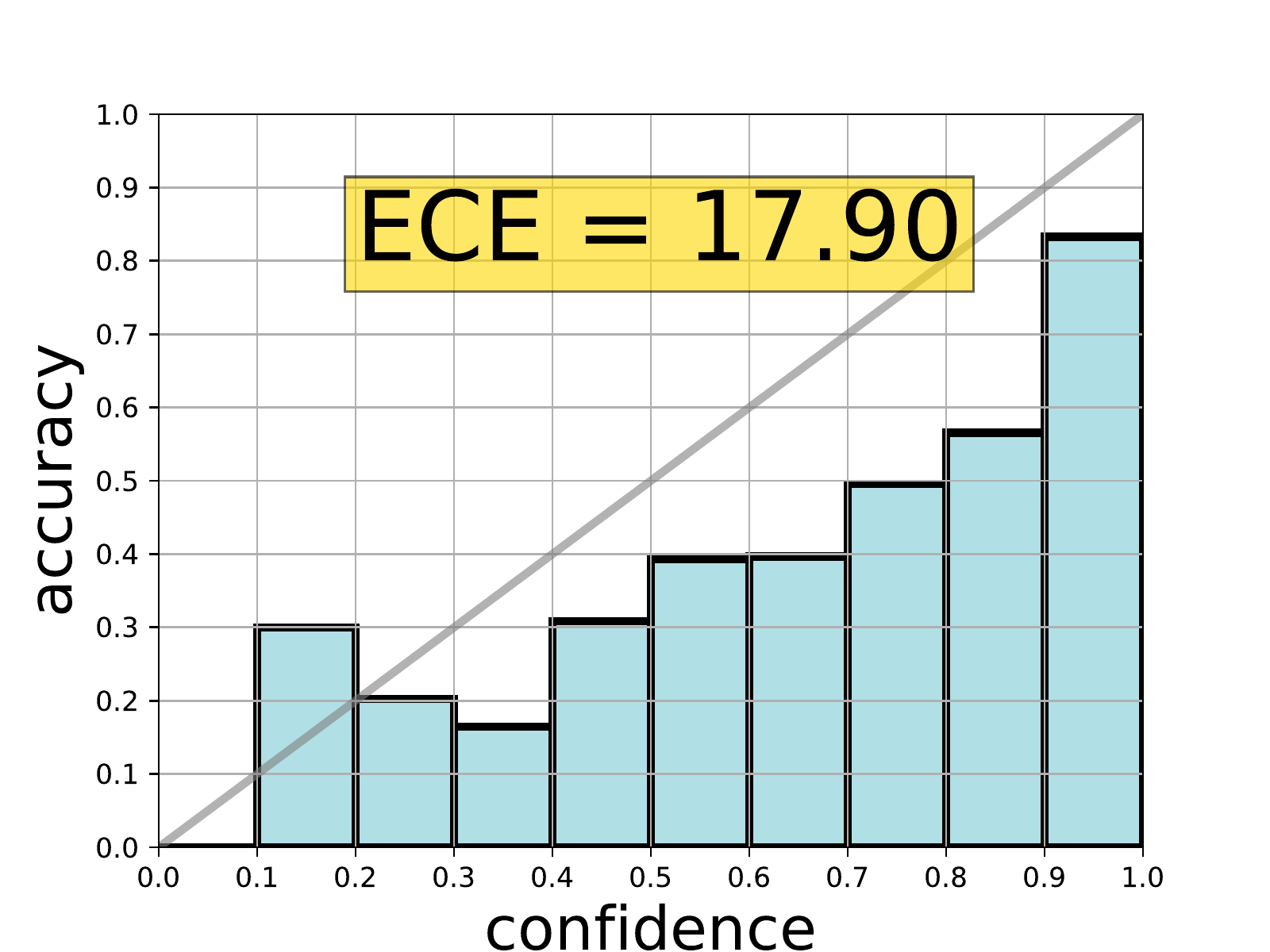}
		\caption{P3D (original), \\ \null \quad all action classes}
		\label{fig:p3d_all}
	\end{subfigure}%
	\begin{subfigure}{.16\textwidth}
		\centering
		\includegraphics[trim={0.1cm 0 1.4cm 0},clip,width=\linewidth]{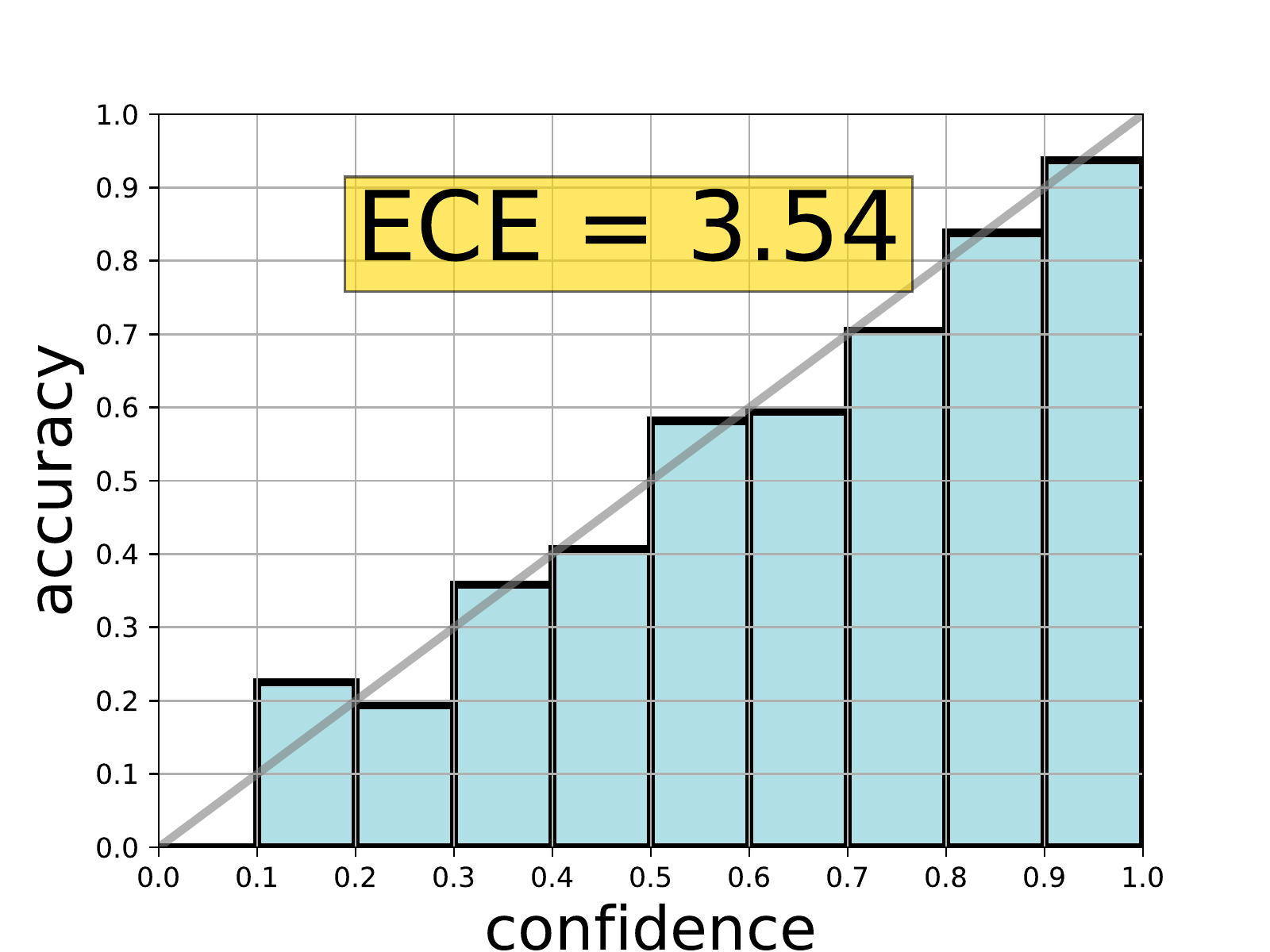}
		\caption{P3D + temp. scaling, all action classes}
		\label{fig:p3d_temp_all}
	\end{subfigure}%
	\begin{subfigure}{.16\textwidth}
		\centering
		\includegraphics[trim={0.1cm 0 1.4cm 0},clip,width=\linewidth]{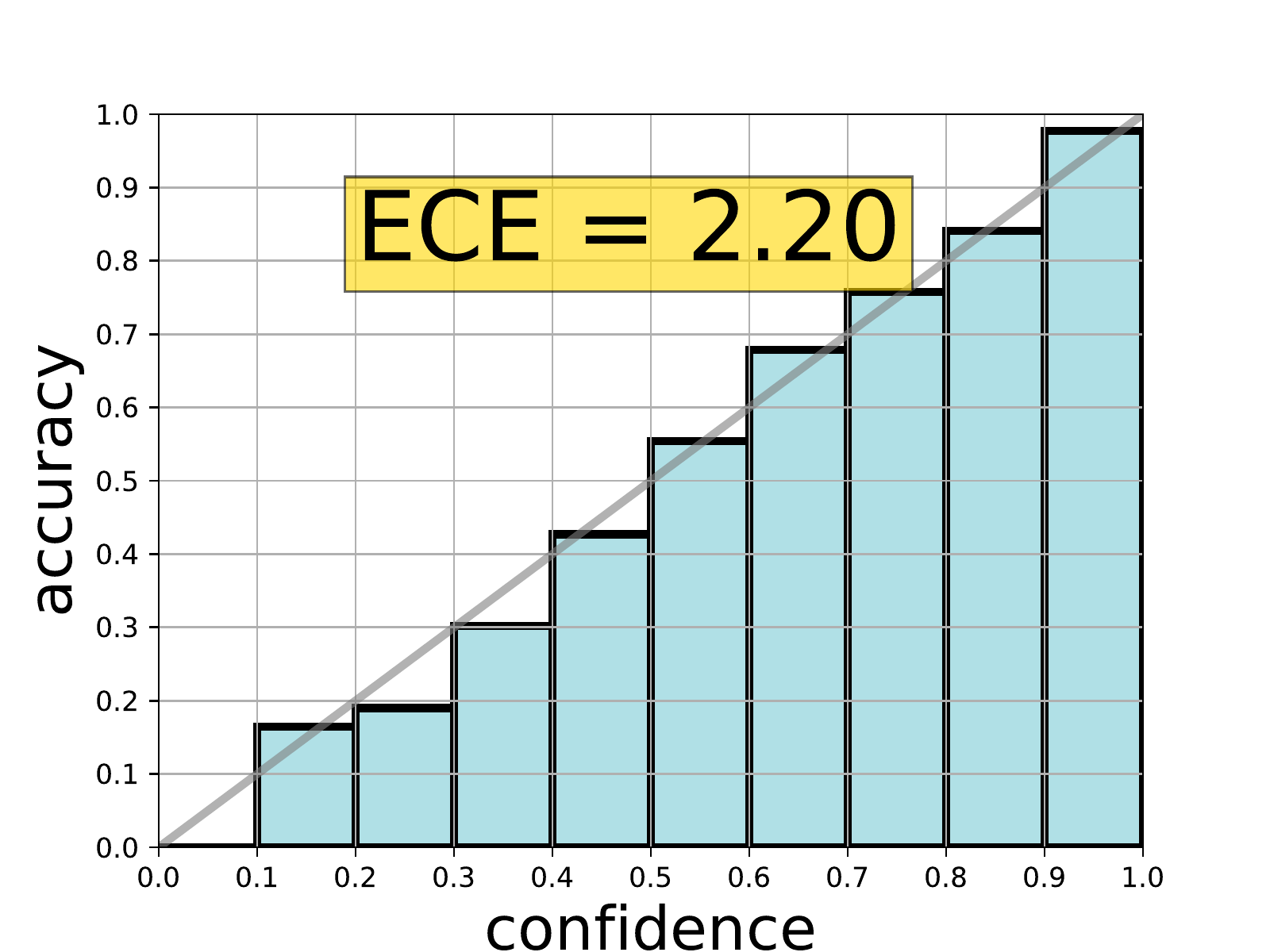}
		\caption{CARING-P3D, \\ \null \quad all action classes}
		\label{fig:p3d_caring_all}
	\end{subfigure}%
	
	\begin{subfigure}{.16\textwidth}
		\centering
		\includegraphics[trim={0.1cm 0 1.4cm 0},clip,width=\linewidth]{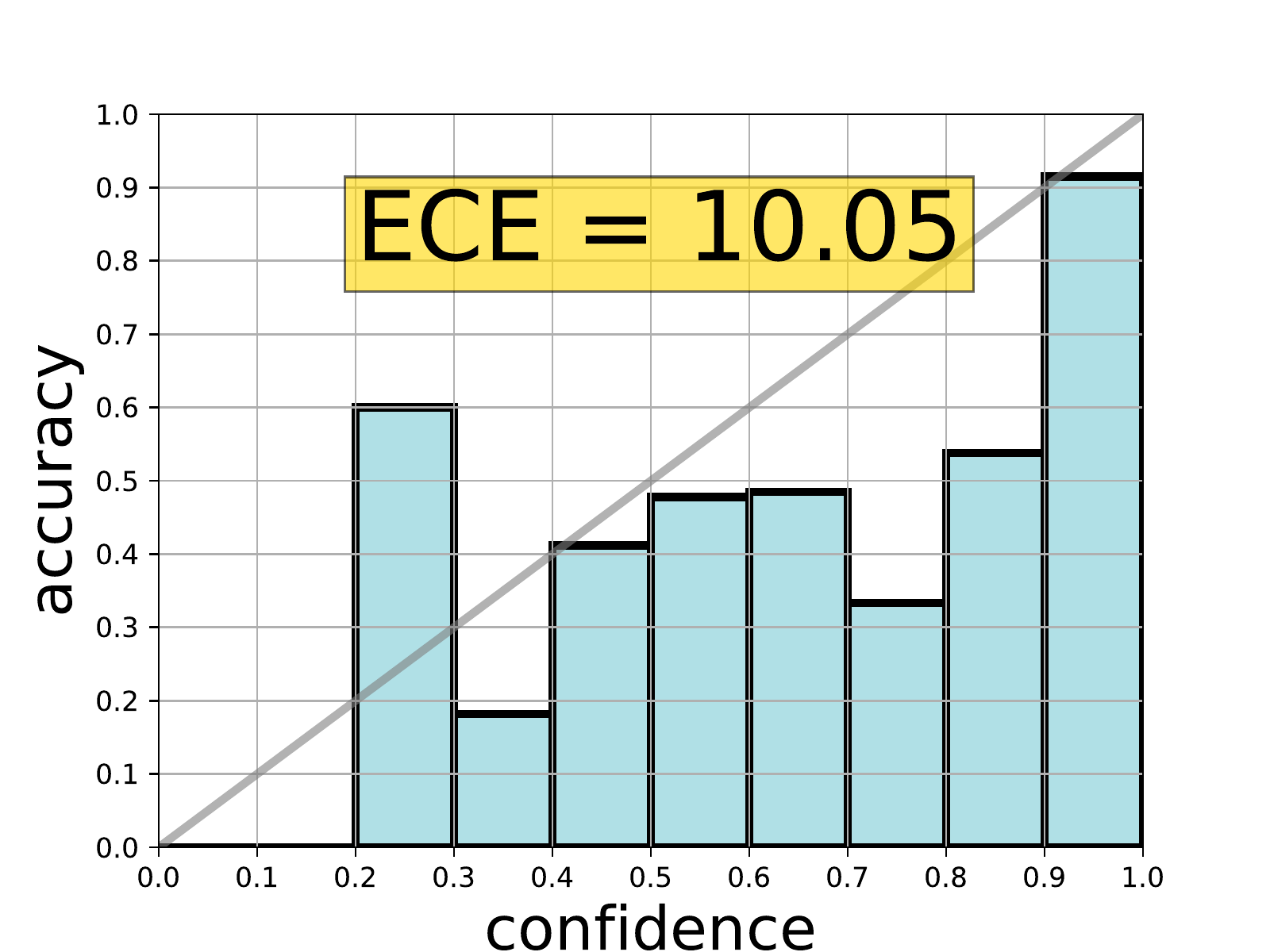}
		\caption{I3D (original), \\ \null \quad common classes}
		\label{fig:i3d_common}
	\end{subfigure}%
	\begin{subfigure}{.16\textwidth}
		\centering
		\includegraphics[trim={0.1cm 0 1.4cm 0},clip,width=\linewidth]{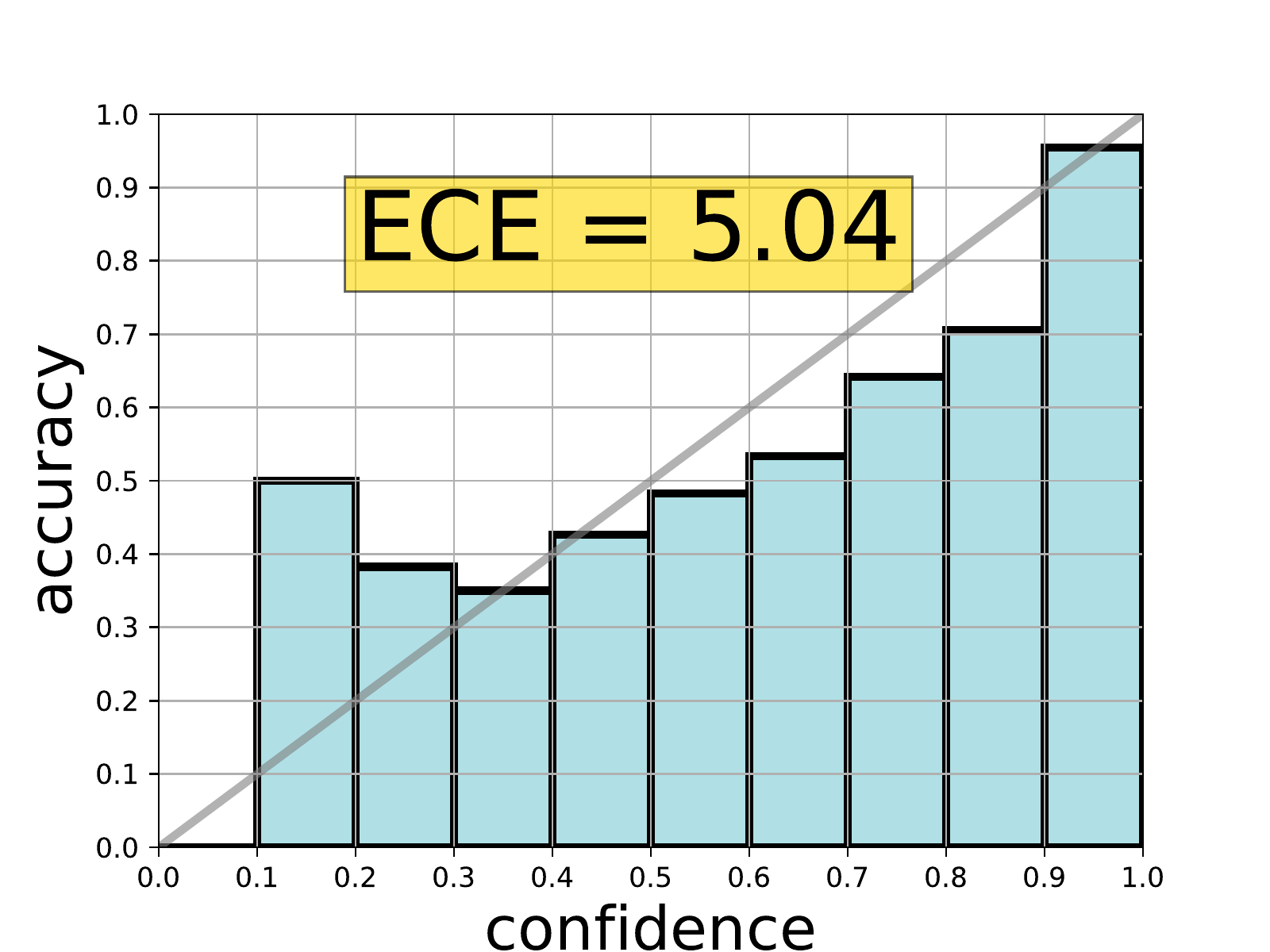}
		\caption{I3D + temp. scaling, common classes}
		\label{fig:i3d_temp_common}
	\end{subfigure}%
	\begin{subfigure}{.16\textwidth}
		\centering
		\includegraphics[trim={0.1cm 0 1.4cm 0},clip,width=\linewidth]{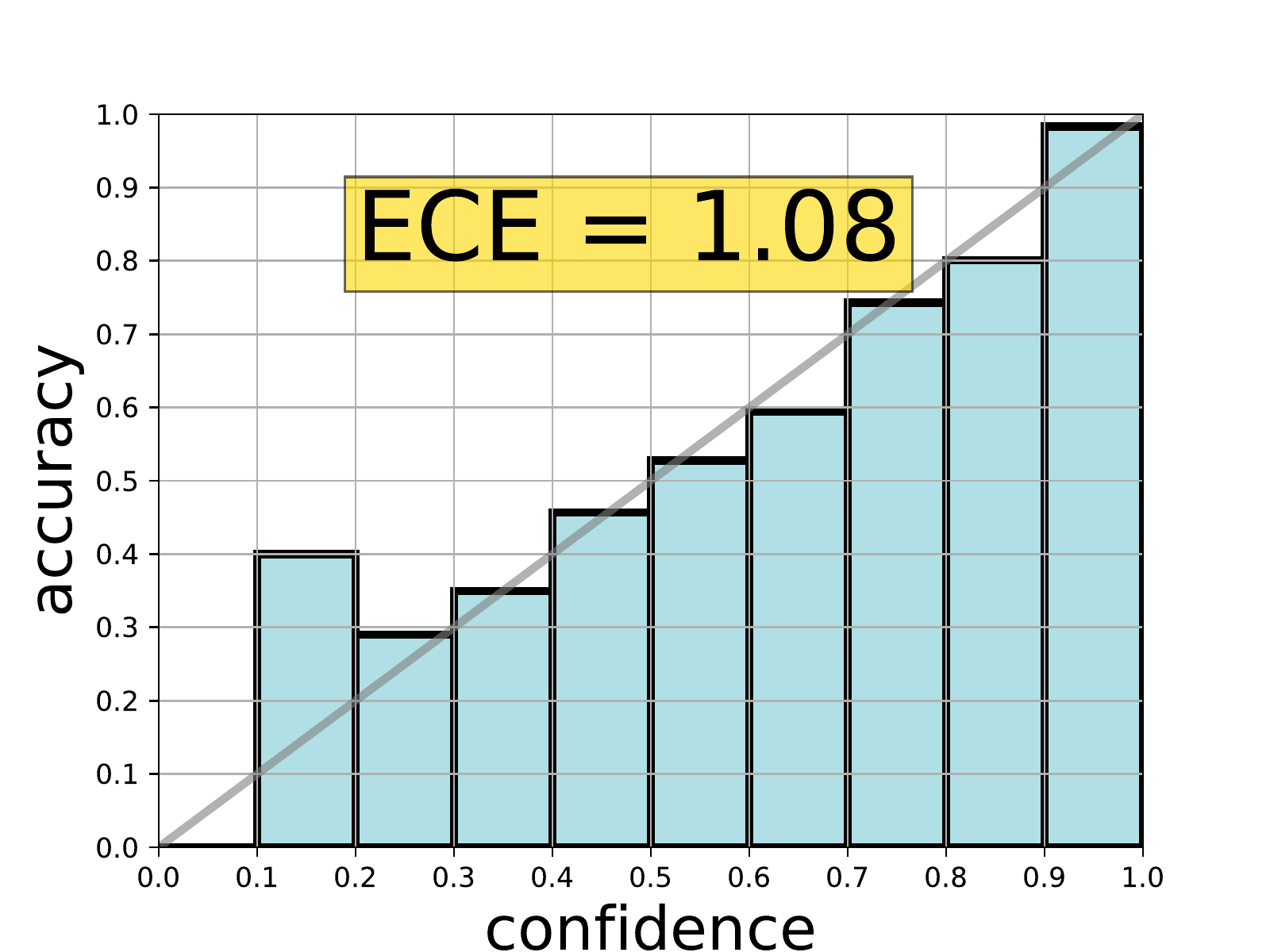}
		\caption{CARING-I3D, \\ \null \quad common classes}
		\label{fig:i3d_caring_common}
	\end{subfigure}%
	\begin{subfigure}{.16\textwidth}
		\centering
		\includegraphics[trim={0.1cm 0 1.4cm 0},clip,width=\linewidth]{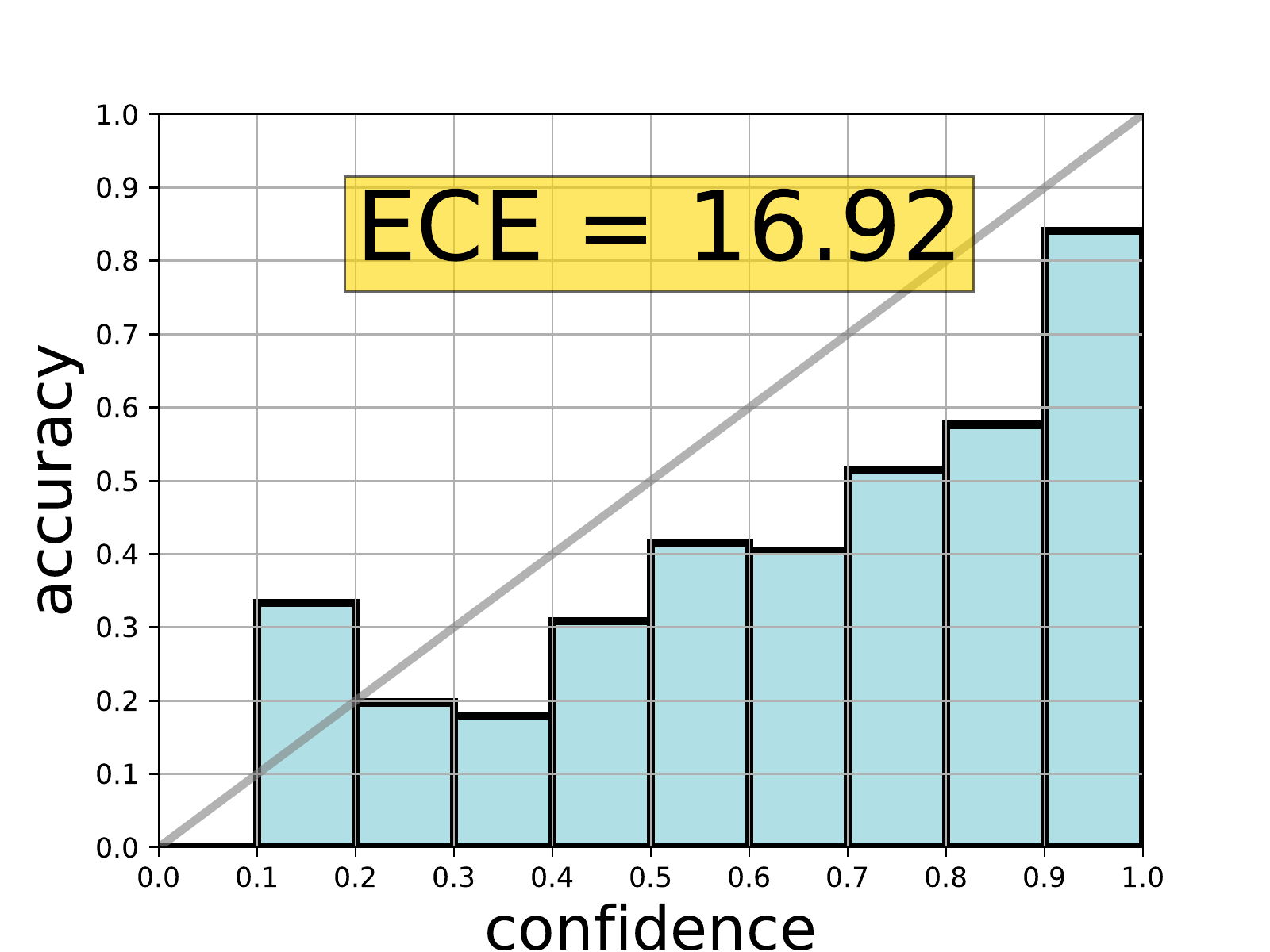}
		\caption{P3D (original), \\ \null \quad common classes}
		\label{fig:p3d_common}
	\end{subfigure}%
	\begin{subfigure}{.16\textwidth}
		\centering
		\includegraphics[trim={0.1cm 0 1.4cm 0},clip,width=\linewidth]{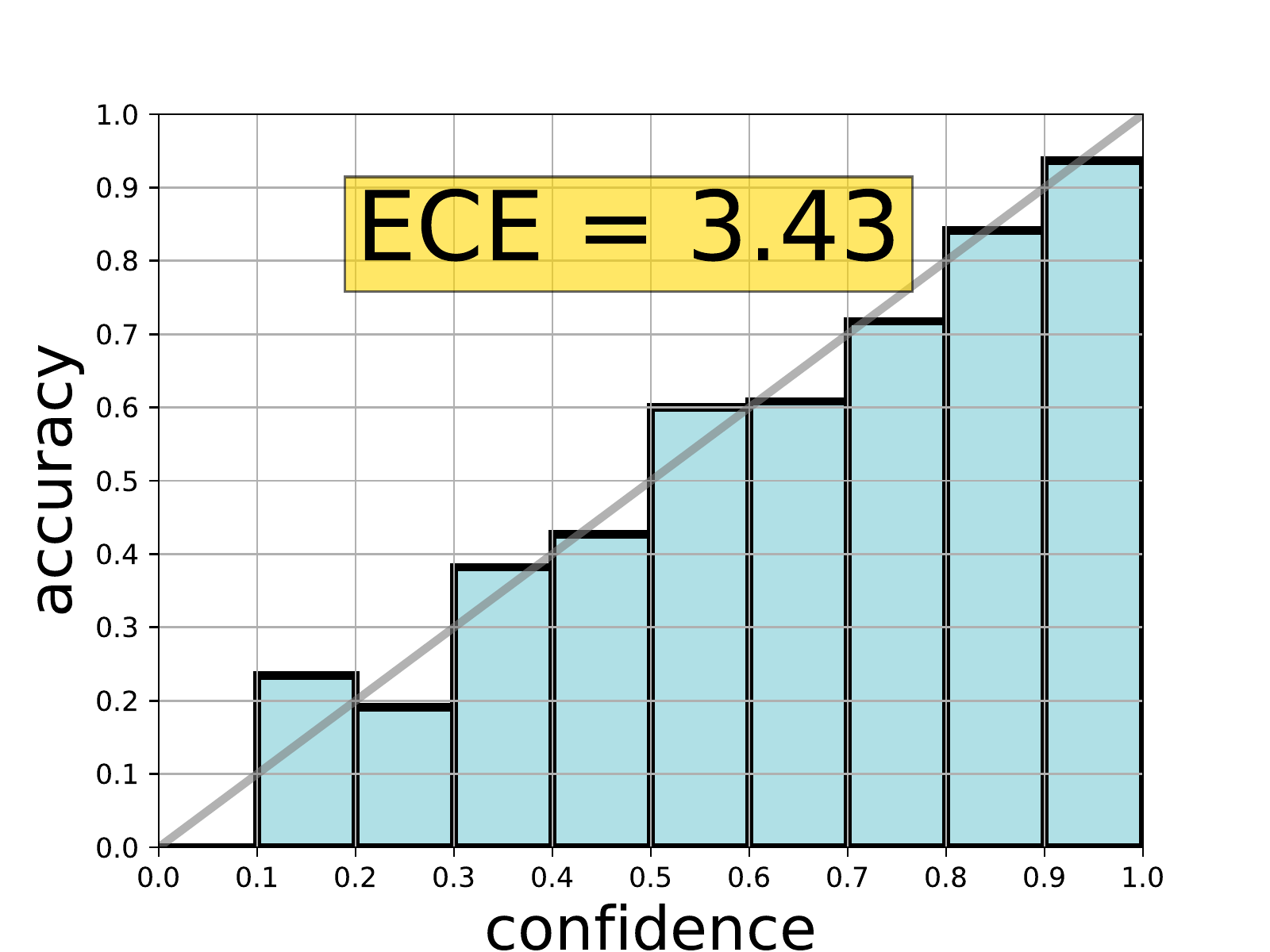}
		\caption{P3D + temp. scaling, common classes}
		\label{fig:p3d_temp_common}
	\end{subfigure}%
	\begin{subfigure}{.16\textwidth}
		\centering
		\includegraphics[trim={0.1cm 0 1.4cm 0},clip,width=\linewidth]{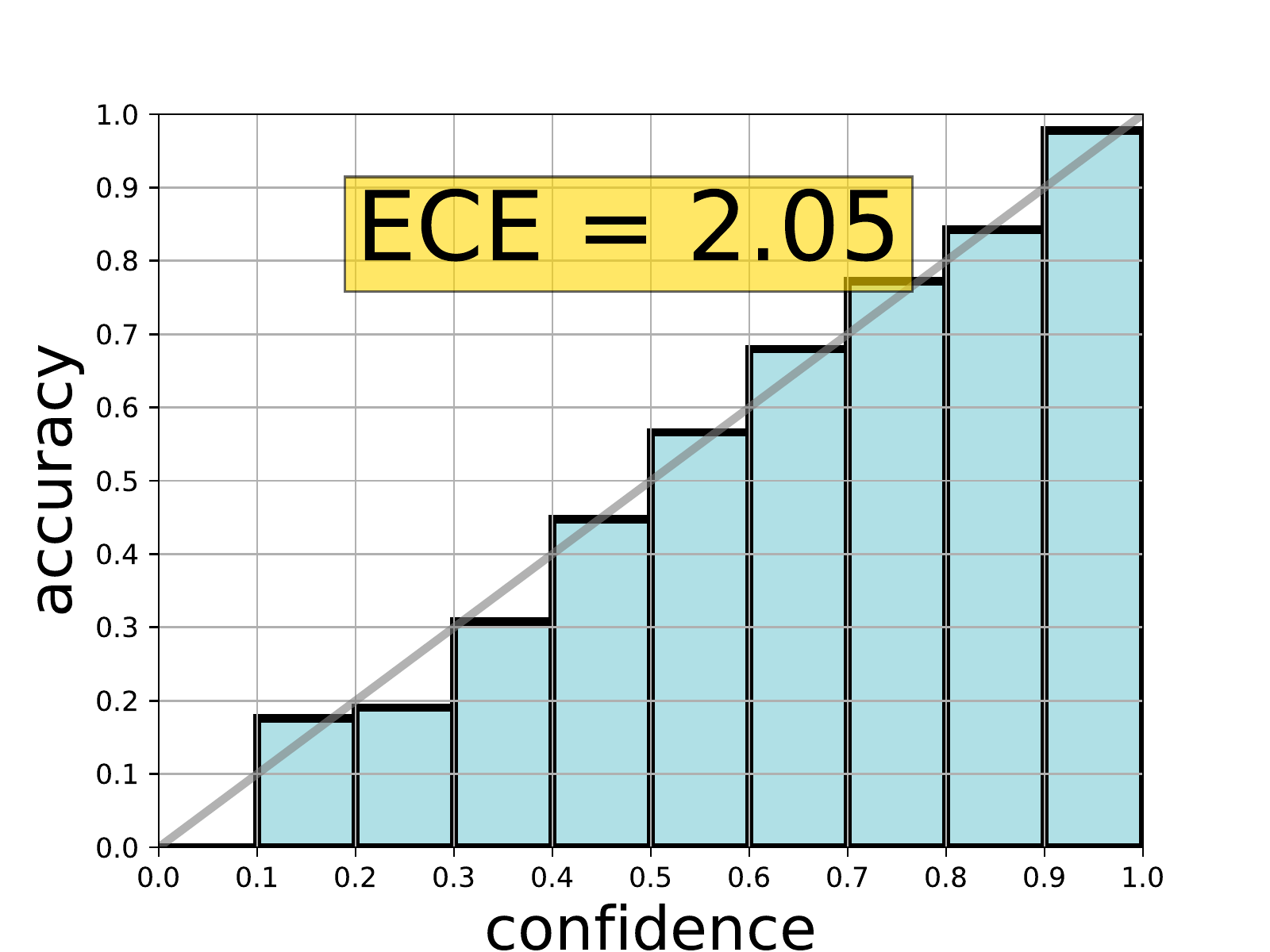}
		\caption{CARING-P3D, \\ \null \quad common classes}
		\label{fig:p3d_caring_common}
	\end{subfigure}%
	
	\begin{subfigure}{.16\textwidth}
		\centering
		\includegraphics[trim={0.1cm 0 1.4cm 0},clip,width=\linewidth]{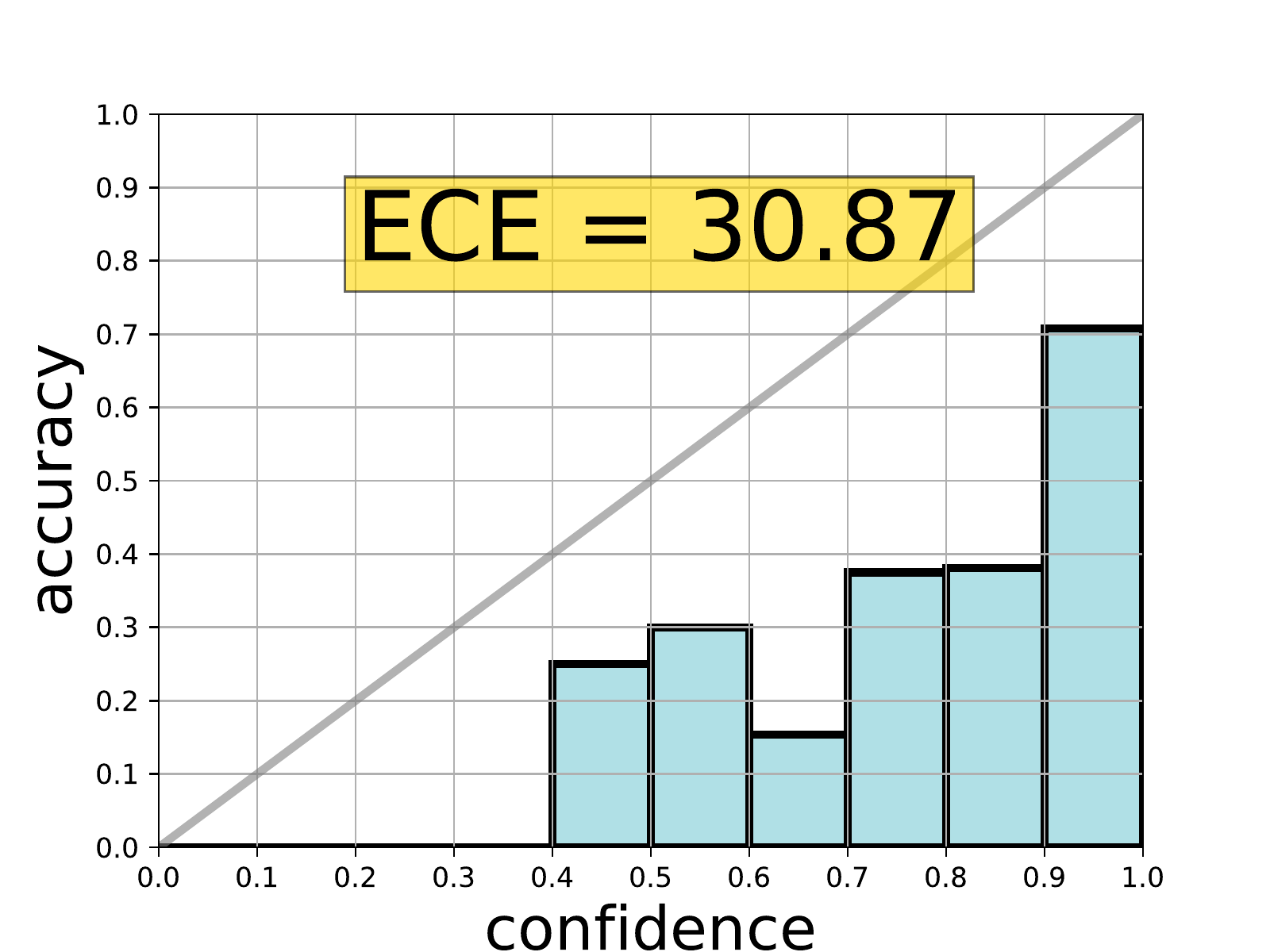}
		\caption{I3D (original), \\ \null \quad rare   classes}
		\label{fig:i3d_rare}
	\end{subfigure}%
	\begin{subfigure}{.16\textwidth}
		\centering
		\includegraphics[trim={0.1cm 0 1.4cm 0},clip,width=\linewidth]{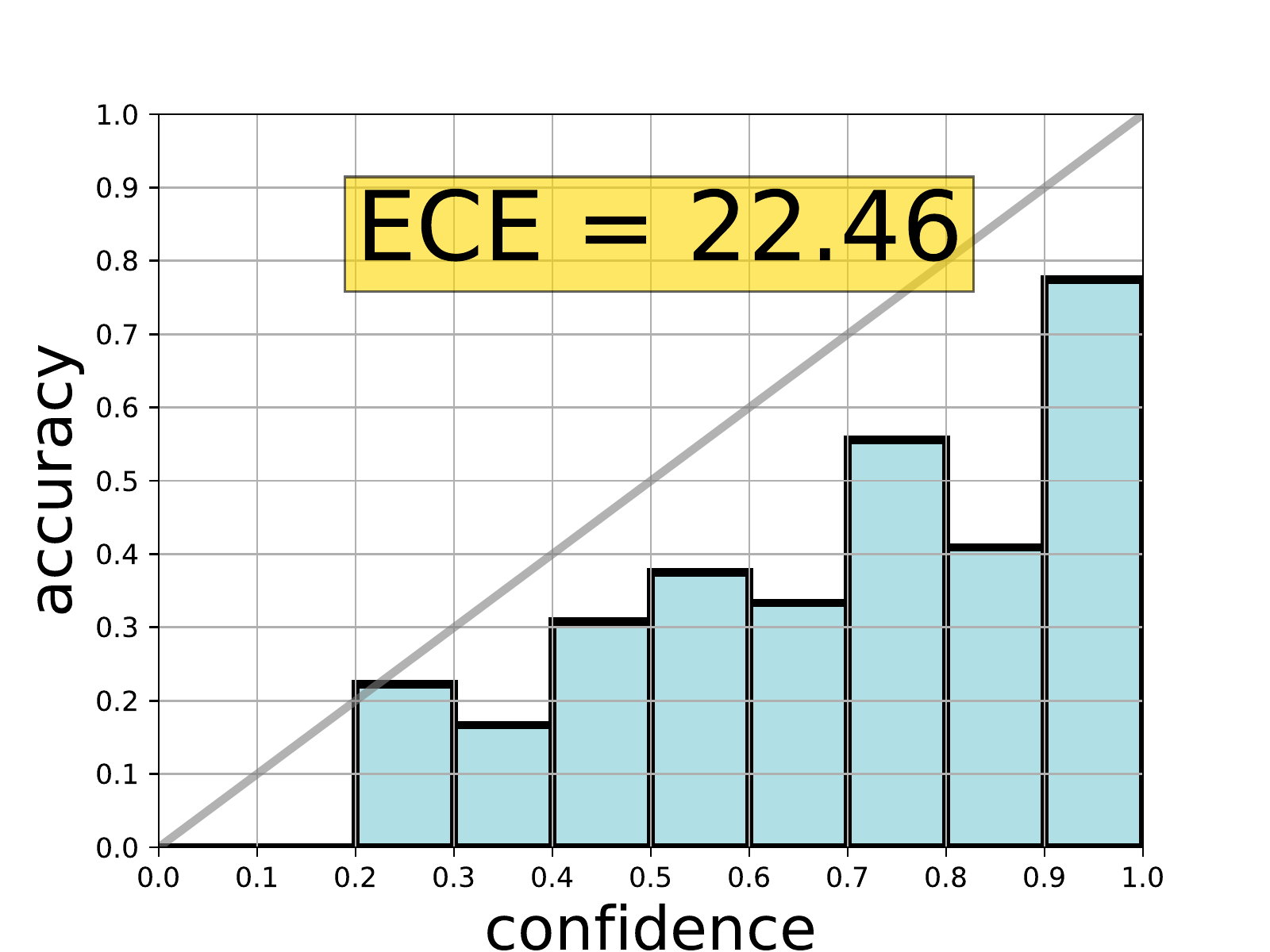}
		\caption{I3D + temp. scaling, rare  classes}
		\label{fig:i3d_temp_rare}
	\end{subfigure}%
	\begin{subfigure}{.16\textwidth}
		\centering
		\includegraphics[trim={0.1cm 0 1.4cm 0},clip,width=\linewidth]{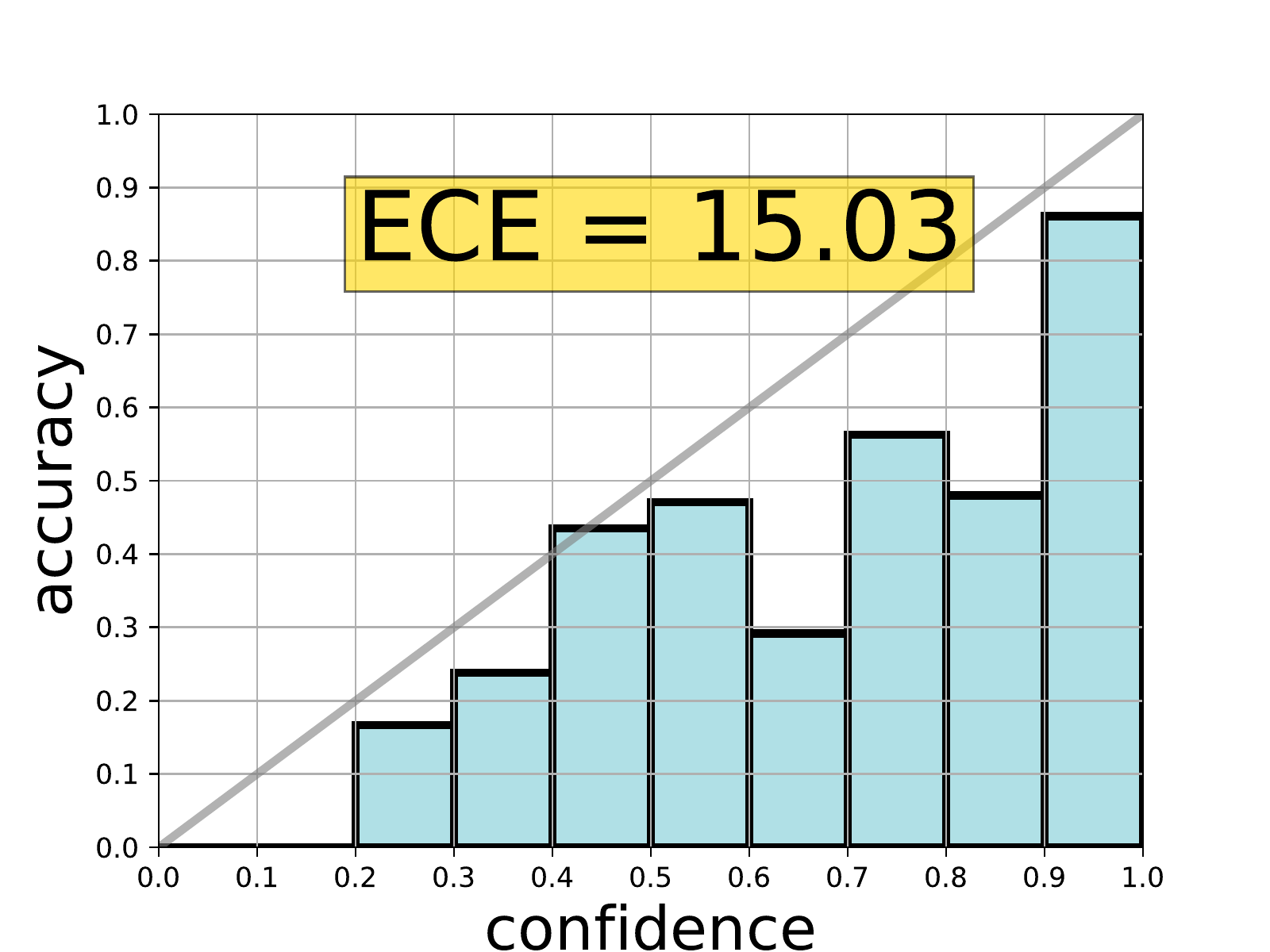}
		\caption{CARING-I3D, \\ \null \quad rare  classes}
		\label{fig:i3d_caring_rare}
	\end{subfigure}%
	\begin{subfigure}{.16\textwidth}
		\centering
		\includegraphics[trim={0.1cm 0 1.4cm 0},clip,width=\linewidth]{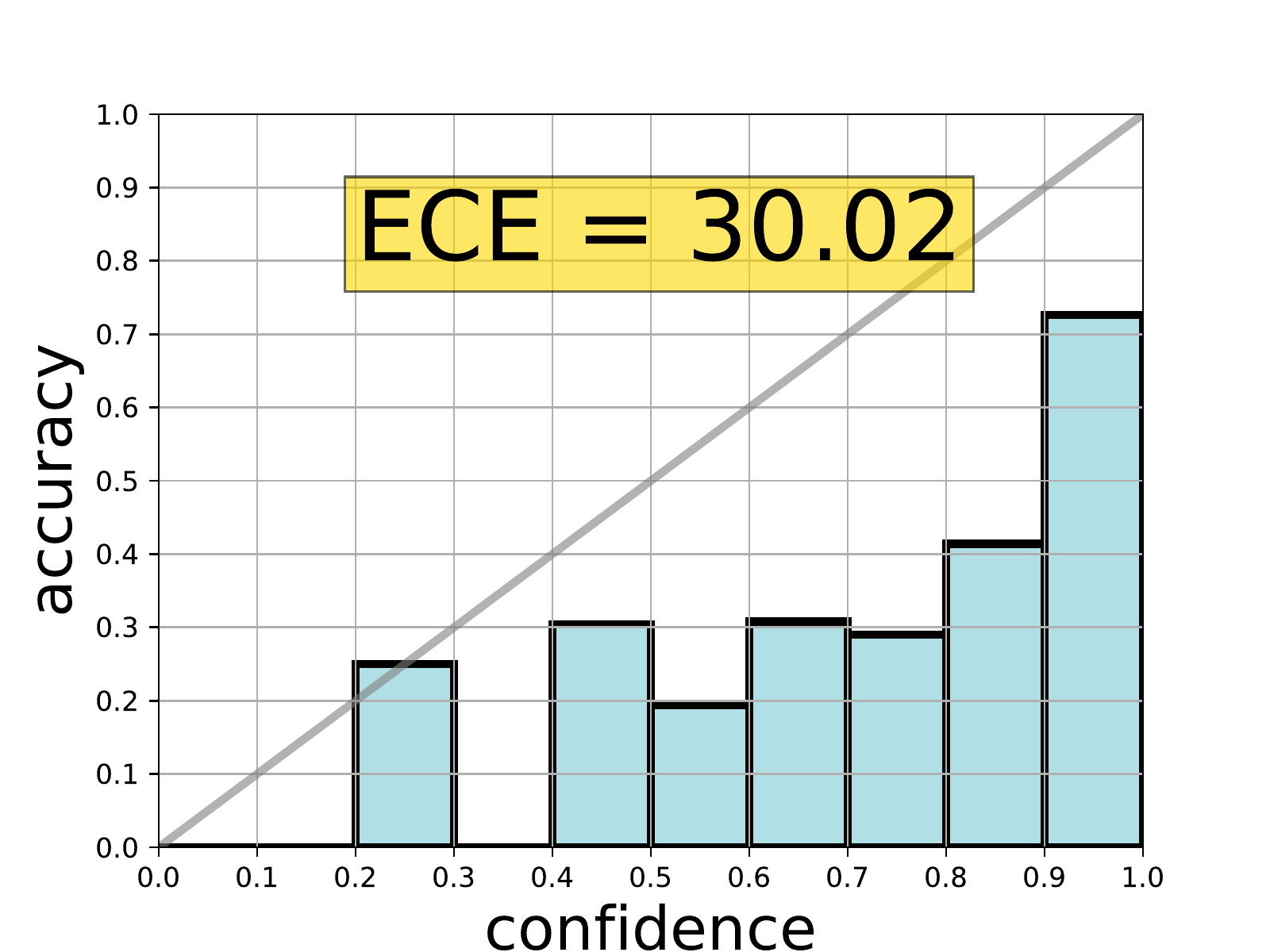}
		\caption{P3D (original), \\ \null \quad rare  classes}
		\label{fig:p3d_rare}
	\end{subfigure}%
	\begin{subfigure}{.16\textwidth}
		\centering
		\includegraphics[trim={0.1cm 0 1.4cm 0},clip,width=\linewidth]{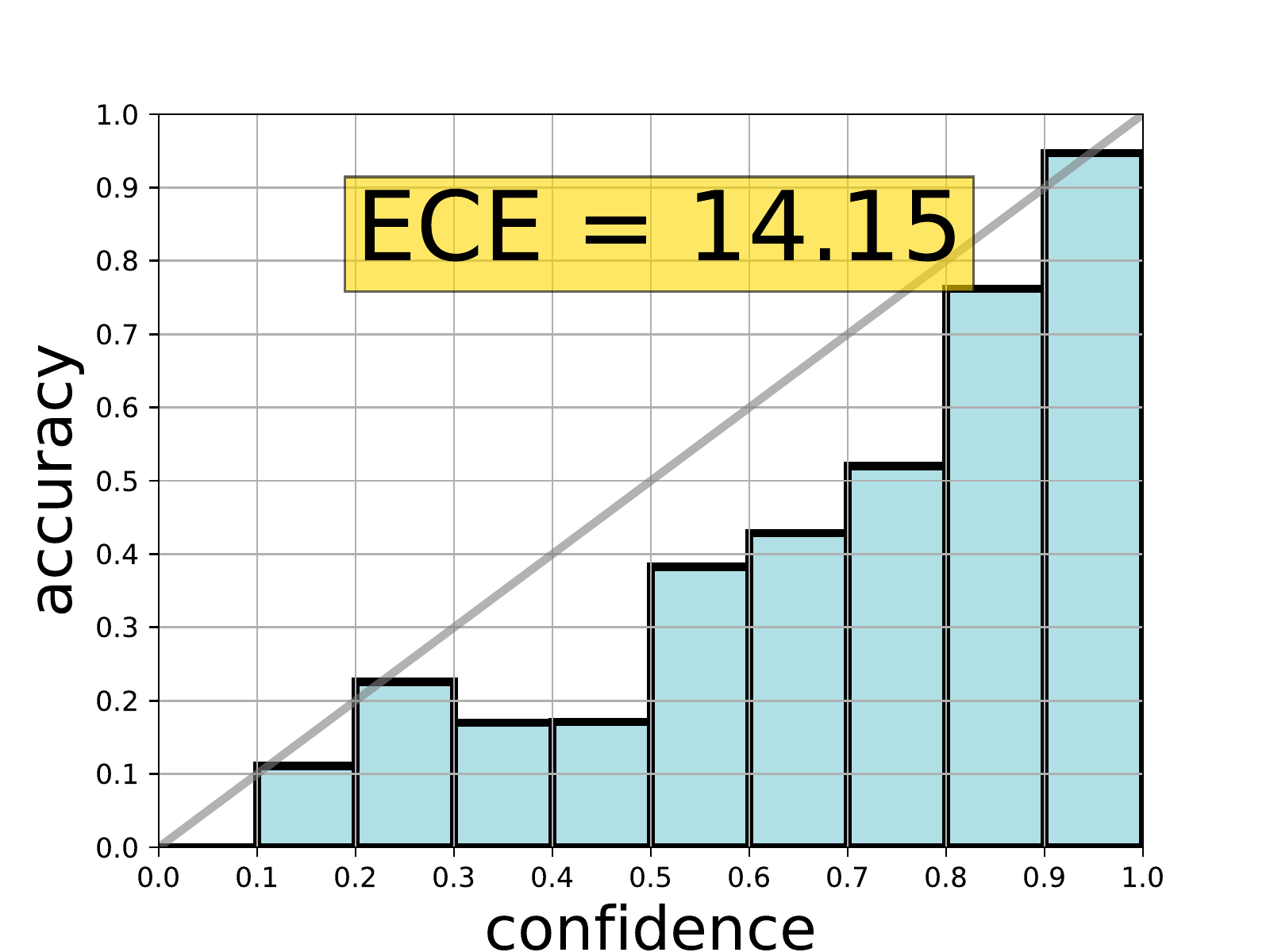}
		\caption{P3D + temp. scaling, rare  classes}
		\label{fig:p3d_temp_rare}
	\end{subfigure}%
	\begin{subfigure}{.16\textwidth}
		\centering
		\includegraphics[trim={0.1cm 0 1.4cm 0},clip,width=\linewidth]{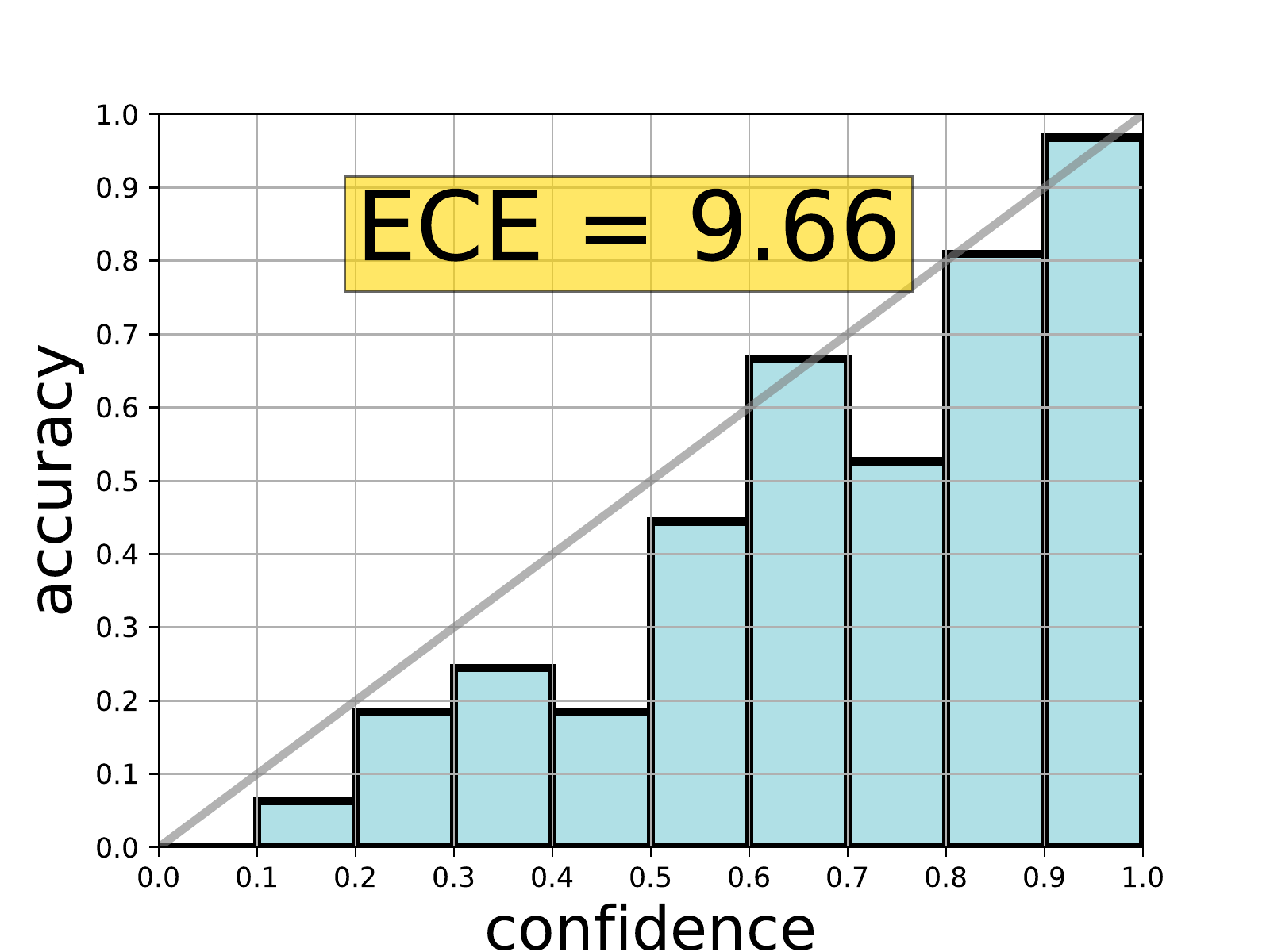}
		\caption{CARING-P3D, \\ \null \quad rare  classes}
		\label{fig:p3d_caring_rare}
	\end{subfigure}%
	\caption{Reliability diagrams of different models reflect the agreement between the confidence values and the empirically measured probability of  correct prediction (results of one Drive\&Act validation split). A model with \emph{perfectly calibrated uncertainty scores would match the diagonal} (a detailed explanation in Section \ref{sec:calib_diag} ). Note, that the ECE values deviate from Table \ref{tbl:reliability_results}, as they visualize a single split, while the final reported results are averaged over all splits.
		While the temperature scaling consistently improves the confidence estimates, our CARING model leads to the lowest calibration error in all settings. } 
	\label{fig:reliability_diag}
	\vspace{-0.1cm}
\end{figure*}

In Figure \ref{fig:reliability_diag}, we visualize the agreement between the predicted model confidence and the empirically measured probability of the correct outcome  via reliability diagrams (explained in Section \ref{sec:definition}) . 
%First, the space of possible probabilities is discretized into $N$ bins (we choose $N=10$). 
%Samples, which predicted confidence lies between 0 and 0.1 falls into first bin, between 0.1 and 0.2 into second bin and so on. 
In case of good estimates, the result will be close to the diagonal line. 
Values above the diagonal are linked to models being overly confident in their prediction, while values below indicate that the model doubts the outcome too much and the accurate prediction probability is higher than assumed. 

First, we discuss the reliability diagrams of the original action recognition networks.
Both P3D and I3D confidence values deviate from the target, with a clear bias towards too optimistic scores  (\ie values are oftentimes below the diagonal in Figures \ref{fig:i3d_all}, \ref{fig:p3d_all}, \ref{fig:i3d_common}, \ref{fig:p3d_common}, \ref{fig:i3d_rare}, \ref{fig:p3d_rare}).
One exception is an above-diagonal peak in the low probability segment for \textit{all} and \textit{common} classes, meaning that in ``easier'' settings,  low confidence examples often turn out to be correct (\ref{fig:i3d_all}, \ref{fig:p3d_all}, \ref{fig:i3d_common}, \ref{fig:p3d_common}). 
%Still, in most probability value segments the prediction seems to fail more often, then the resulting Softmax score indicates.
In the ``harder'' setting of \emph{rare} activities (Figure \ref{fig:i3d_rare}, \ref{fig:p3d_rare}), the bias towards too high  probabilities is present for all values.

We see a clear positive impact of temperature scaling (Figures \ref{fig:i3d_temp_all}, \ref{fig:p3d_temp_all}, 
\ref{fig:i3d_temp_common}, \ref{fig:p3d_temp_common},  \ref{fig:i3d_temp_rare}, \ref{fig:p3d_temp_rare}) and our CARING model (Figures \ref{fig:i3d_caring_all}, \ref{fig:p3d_caring_all}, 
\ref{fig:i3d_caring_common},  \ref{fig:p3d_caring_common},  \ref{fig:i3d_caring_rare}, \ref{fig:p3d_caring_rare}).
%While temperature scaling clearly improves calibration of model confidence, reliability diagrams of the 
CARING models outperform other approaches in all settings and lead to almost perfect reliability diagrams for \textit{all} and \textit{common} classes.
%C3D therefore is well suited for coarse classification 
Still, both temperature scaling and CARING methods have issues with rare classes, with model confidence still being too high, marking an important direction for future research.
%While this is unsurprising, as deep models are notably bad in learning from few examples, this setting is important and has room for  improvement in the future.

Note, that ECE might be in a slight disarray with the visual reliability diagram representation, as the metric weighs the misalignment in each bin by the amount of data-points in it, while the reliability diagrams do not reflect such frequency distribution. 
For example, while the CARING-I3D model in Figure~\ref{fig:i3d_caring_common} slightly exceeds the target diagonal, it has lower expected calibration error than CARING-P3D which seems to produce nearly perfect results in Figure~\ref{fig:p3d_caring_common}. 
As there are only very few examples in the low-confidence bin, they are overshadowed by smaller differences in the high-confidence bins, which contribute much more as they have more samples.
%From the far left diagram, we see that theuncalibrated ResNet tends to be overconfident in its pre-dictions.  We then can observe the effects of temperaturescaling (middle left), histogram binning (middle right), andisotonic regression (far right) on calibration. All three dis-played methods produce much better confidence estimates.Of the three methods, temperature scaling most closely re-covers the desired diagonal function.  Each of the bins arewell calibrated, which is remarkable given that all the prob-abilities were modified by only a single parameter.  We in-clude reliability diagrams for other datasets in Section S4.

%\subsection{Reliability Analysis of the Individual Activity Classes}

%!TEX root = ../root.tex
\section{Conclusion}

Activity understanding opens doors for new ways of human-machine
interaction but requires models that can identify uncertain situations.
%Beyond assigning the correct activity class, an activity recognition model should be able to understand, how confident it is in its predictions.  
We go beyond the traditional goal of high top-1 accuracy and make the first step towards activity recognition models capable of \emph{identifying their misclassifications}.
%To this intent, we  measure reliability of the confidence score and evaluate it for two prominent action recognition architectures, revealing, that the raw Softmax values of such models do not reflect the probability of correct prediction well.
We measure the \emph{reliability of model confidence} and evaluate it for two prominent action recognition models, revealing, that the raw Softmax values of such networks do not reflect the probability of a correct prediction well.
We further implement two strategies for learning to convert poorly calibrated confidence values into realistic uncertainty estimates.
First, we combine the native action recognition models with the off-the-shelf temperature scaling~\cite{guo2017calibration} approach which divides the network logits by a single learned scalar. 
We then introduce a new approach which learns to produce individual input-guided temperature values  dependent on the input representation through an additional calibration network.
We show in a thorough evaluation, that our model consistently outperforms the temperature scaling method and native activity recognition networks in producing realistic confidence estimates.
%While the proposed methods clearly lead to more realistic uncertainty estimates, decreasing the expected calibration error to xx\% on the Drive\&Act dataset, there is room for improvement in case of few training examples. 
%The experiment results hold great promise for uncertainty-aware models, a crucial step towards real-life applications of activity recognition algorithms.

%\balance
\bibliography{egbib}{}
\bibliographystyle{plain}
\end{document}